%% file: main.tex
\documentclass{article} 
\usepackage{iclr2027_conference,times}

\input{math_commands.tex}

\usepackage{url}
\usepackage{booktabs}
\usepackage{tabularx}
\usepackage{amssymb}
\usepackage{amsthm}
\usepackage{enumitem}
\newtheorem{theorem}{Theorem}
\newtheorem{proposition}[theorem]{Proposition}
\newtheorem{corollary}[theorem]{Corollary}
\usepackage{multirow}
\usepackage{graphicx}
\usepackage{capt-of}
\usepackage{float}
\floatstyle{ruled}
\newfloat{algorithm}{tbp}{loa}
\floatname{algorithm}{Algorithm}
\floatstyle{plain}
\usepackage{wrapfig}
\usepackage{needspace}
\usepackage{placeins}
\usepackage{longtable}

\usepackage{tcolorbox}
\usepackage{xcolor}
\usepackage{microtype}
\usepackage[hidelinks]{hyperref}

\definecolor{resultbest}{RGB}{0,120,62}
\definecolor{resultworst}{RGB}{190,35,35}

\newcommand{\yichen}[1]{\textcolor{black}{#1}}

\newcommand{\rev}[1]{\textcolor{black}{#1}} 

\title{FEAST: Fair and Efficient Adaptive Sampling for Task-Free On-Policy Learning}

\author{
  Yongxi Zhu\textmd{\textsuperscript{1}}{\footnotemark[1]}\textmd{,} 
  ~Ziyi Wang\textmd{\textsuperscript{1}}\footnotemark[1]\textmd{,}
  ~Yichen Wu\textmd{\textsuperscript{2}}\textmd{,}
  ~Ruite Lu\textmd{\textsuperscript{1}}\textmd{,}
  ~Chang Liao\textmd{\textsuperscript{3}}\textmd{,}
  ~Haipeng Li\textmd{\textsuperscript{4}}\textmd{,}
  ~Ying Wei\textmd{\textsuperscript{{1}}}\footnotemark[2] \\
  \textsuperscript{1}Zhejiang University, \textsuperscript{2}Harvard University, \textsuperscript{3}Nanyang Technological University, \\ \textsuperscript{4}The Hong Kong University of Science and Technology
}

\newif\ifisfinalcopy
\isfinalcopytrue

\ifisfinalcopy
    \iclrfinalcopy
\fi

\begin{document}
\raggedbottom
\setlength{\parskip}{4.5pt}

\maketitle

\ifisfinalcopy
    \footnotetext[1]{Equal contribution.}
    \footnotetext[2]{Corresponding author.}
\fi

\input{sec/0-abstract}
\input{sec/1-introduction}
\input{sec/2-relatedWork}
\input{sec/3-qel}
\input{sec/4-feast}
\input{sec/5-experiments}
\input{sec/6-conclusion}

\bibliography{main}
\bibliographystyle{iclr2027_conference}

\appendix
\input{sec/appendix}

\end{document}

%% file: math_commands.tex
\usepackage{amsmath,amsfonts,bm}

\def\eqref#1{equation~\ref{#1}}

\def\1{\bm{1}}

\DeclareMathAlphabet{\mathsfit}{\encodingdefault}{\sfdefault}{m}{sl}
\SetMathAlphabet{\mathsfit}{bold}{\encodingdefault}{\sfdefault}{bx}{n}



%% file: sec/0-abstract.tex
\begin{abstract}

\yichen{Language agents trained from self-generated experience must allocate a limited rollout budget across heterogeneous capabilities.
A common strategy is to prioritize capabilities with strong current learning signals, but in on-policy learning these signals are endogenous: past allocation changes the policy and therefore the evidence available for future allocation. This can create a feedback loop in which under-trained capabilities receive increasingly fewer opportunities to improve.
We formalize this effect through the Quota-Efficiency Law (QEL), which shows that allocation history can change learning efficiency even under matched capability-specific exposure. Motivated by this, we introduce FEAST, a task-free adaptive sampling framework that separates opportunity allocation from compute allocation. FEAST dynamically discovers capability clusters while preserving their learning histories, maintains continued sampling opportunities across clusters, and reserves a rollout budget for each selected cluster. Within each budget, it progressively concentrates computation on prompts with informative reward variation. Across GURU-18K and SMC, FEAST improves average performance over competitive adaptive sampling baselines under matched rollout budgets, while better preserving learning progress on weaker capabilities. Ablation studies further support the complementary roles of cross-cluster opportunity protection and within-cluster selective allocation. Code will be released upon acceptance. }

\end{abstract}

%% file: sec/1-introduction.tex
\section{Introduction}
\label{sec:introduction}

\input{figures/guru8b_teaser}


An agent that improves through reinforcement learning must \yichen{decide where to spend its training experience}. Reinforcement learning with verifiable rewards (RLVR) enables \yichen{such} self-evolution by \yichen{allowing} a language model to generate attempts, receive outcome feedback, and update its policy~\citep{shao2024deepseekmath,guo2025deepseekr1}. As \yichen{RLVR} expands from mathematics to heterogeneous reasoning domains~\citep{cheng2025guru,ma2025general,akter2026nemotron}, \yichen{experience allocation becomes increasingly important:} a fixed rollout budget must serve capabilities that differ in difficulty, current proficiency, and trainability. Moreover, \yichen{predefined domain labels may not reflect the capabilities that matter to the current learner. The resulting challenge is to allocate experience where it is most needed without relying on a fixed task taxonomy.}


Adaptive curricula address this challenge by estimating learning utility and \yichen{prioritizing} promising data. Learning progress, policy advantages, and cross-domain transfer all provide useful guidance~\citep{graves2017automated,wang2025dump,chen2026sec,yang2026tac}. In on-policy learning, however, utility is not an intrinsic property of an example or capability: \yichen{allocation decisions change the policy, which in turn changes the outcomes available for future learning.} Recent work has begun to recognize this endogenous nonstationarity~\citep{mckenzie2026bmc}. \yichen{As a result, a weak capability may show little immediate learning signal simply because it has received too few opportunities to improve. Prioritizing only the strongest observed signals can then create a feedback loop in which under-allocation produces weak evidence, and weak evidence justifies further under-allocation.}

We formalize this \yichen{phenomenon} through the {Quota--Efficiency Law} (QEL), which compares allocation paths at matched exposure to \yichen{a target capability.} \yichen{Consider two schedules that train the same capability at different frequencies. We allow the sparser schedule to run longer until both schedules have generated the same number of samples for that capability. Updates on other capabilities move the policy before later target-capability samples are collected, affecting both its current performance and the quality of its subsequent self-generated experience. QEL therefore captures a path-dependent change in learning efficiency, rather than the trivial effect of receiving fewer samples at the same training step.} This observation motivates \emph{learning-opportunity fairness}: weak current evidence should not permanently exclude a capability from future training. \yichen{This notion does not require equal quotas. Instead, it requires continued opportunities to revisit allocation decisions as the learner evolves. At the same time, protecting a capability does not imply spending computation uniformly on all of its prompts. Some prompts are already saturated, some remain beyond the learner's reach, and others exhibit informative variation in outcomes~\citep{yu2025dapo,jiang2026vigor}. Effective allocation must therefore operate at two levels: deciding which capabilities receive learning opportunities and, within each, where additional computation is most useful.}

We introduce \textbf{FEAST}, a \textbf{F}air and \textbf{E}fficient \textbf{A}daptive \textbf{S}ampling framework for \textbf{T}ask-free agent self-evolution. \yichen{FEAST combines \emph{fairness across capability clusters} with \emph{efficiency within each cluster}. It discovers capability clusters from the learner's representations and learning dynamics while maintaining fine-grained learning histories as those clusters evolve. Across eligible clusters, adaptive priorities favor productive learning opportunities, while explicit protection ensures that low-scoring clusters continue to receive fresh evidence. Each selected cluster is assigned its own rollout budget. Within that budget, FEAST first probes prompts broadly and then progressively concentrates computation on those with informative outcome variation. Because prompts compete only within their cluster's reserved budget, high utility elsewhere cannot eliminate that cluster's opportunity to learn. Figure~\ref{fig:guru8b-teaser} previews FEAST's performance on GURU-18K. Our contributions are threefold:}
\begin{itemize}[leftmargin=3mm, itemsep=-1mm, topsep=-0.5 mm]
\item \rev{We identify allocation history as a factor in learning efficiency. QEL shows that equal training on a capability can yield different gains depending on the preceding allocation path, which motivates learning-opportunity fairness.}
\item \rev{We introduce FEAST, a task-free framework that turns this principle into an allocation rule. It protects learning opportunities across evolving capability clusters and concentrates rollouts within each reserved budget, using persistent learning histories to guide both levels.}
\item We demonstrate improvements over state-of-the-art sampling baselines under matched training and rollout budgets, both in aggregate performance and learning gains on initially weaker capabilities. Together, the ablations and observed training dynamics support protecting learning opportunities and using rollouts selectively within each cluster\rev{, confirming the contribution of each component}.
\end{itemize}

%% file: figures/guru8b_teaser.tex
\begin{wrapfigure}{r}{0.46\textwidth}
\vspace{-7mm}
\centering
\includegraphics[width=0.85\linewidth]{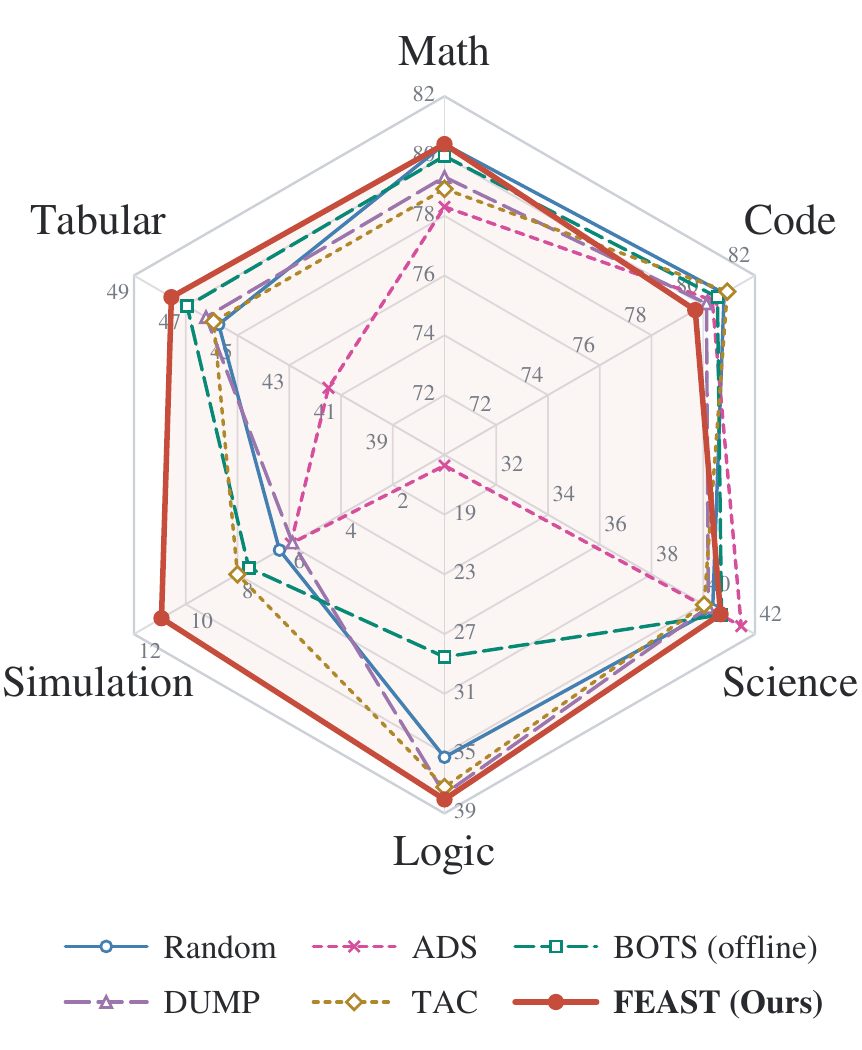}
\vspace{-1mm}
\caption{GURU-18K~\citep{cheng2025guru} results with Qwen3-8B-Base after 100 updates.}
\label{fig:guru8b-teaser}
\end{wrapfigure}

%% file: sec/2-relatedWork.tex
\section{Related Work}
\label{sec:related-work}

\subsection{Multi-Domain Reinforcement Learning}

Reinforcement learning with verifiable rewards increasingly spans heterogeneous reasoning domains. GURU provides multi-domain data and verifiers~\citep{cheng2025guru}, while General-Reasoner broadens answer verification with a generative verifier~\citep{ma2025general}. Nemotron-CrossThink studies data construction and blending beyond mathematical reasoning~\citep{akter2026nemotron}. WildSci extends verifiable training to scientific questions derived from research literature~\citep{liu2026wildsci}. Together, these complementary efforts expand the experience available for RL and highlight the importance of its composition.
Multi-domain data construction determines the available training pool, while an online allocation rule determines which capabilities receive prompts at each update. Under a fixed generation budget, these opportunities compete, and the resulting updates change the policy that produces future attempts and reward signals. Dataset labels also need not match the relevant allocation units: domains can exercise overlapping skills, while prompts within a domain can yield different learning signals for the current policy. FEAST addresses this allocation problem by discovering capability groups and revising their priorities from learning feedback within each capability. It can be applied to diverse, verifiable training data without assuming a fixed domain taxonomy.

\subsection{Adaptive Data Sampling}
\noindent\textbf{Fair and balanced sampling.} Group-level curricula address uneven learning by adapting data mixtures using learning feedback~\citep{graves2017automated}. Skill-It accounts for dependencies among skills~\citep{chen2023skillit}. DUMP combines policy-advantage feedback with UCB exploration~\citep{wang2025dump}, while TAC combines cross-domain transfer estimates with exploratory sampling~\citep{yang2026tac}. SEC updates category priorities through temporal-difference learning~\citep{chen2026sec}, while BMC models how training changes utility over a latent task geometry~\citep{mckenzie2026bmc}. DUMP's UCB bonus is larger for distributions with fewer samples~\citep{wang2025dump}, while Fair-MAB requires each arm to receive a prespecified minimum fraction of pulls~\citep{patil2021fairmab}. DRATS focuses on the largest shortfall from a target return while enforcing a minimum task-sampling probability~\citep{corrado2026drats}. These approaches reflect different criteria for prioritizing training data. Advantage-based selection favors data distributions that currently provide strong training signals. Transfer-aware selection can prioritize a domain whose updates benefit other domains, even when its local learnability is modest.

\input{tables/related_method}

\noindent\textbf{Efficient sampling and rollout allocation.} Efficiency also depends on which examples are selected and how much computation each receives. In supervised fine-tuning, ClusterUCB reduces gradient-based data-selection cost through clustering~\citep{wang2025clusterucb}. For RL, SPaCe uses semantic--difficulty clusters for curriculum sampling~\citep{do2026space}, ADS combines semantic-cluster scheduling with difficulty-aware prompt selection~\citep{xu2026ads}, and ASTOR coordinates task budgets and prompt priorities~\citep{chen2026astor}. BOTS's interpolation estimator combines current-batch outcomes with reference-model scores to estimate difficulty for other prompts~\citep{shen2026bots}. Collecting missing scores incurs a one-time rollout cost that can be amortized across training runs. At the generation level, DAPO filters uniform-outcome groups and refills the batch~\citep{yu2025dapo}, while ODF filters for intermediate-difficulty prompts~\citep{bae2026odf}. GRESO uses reward history to skip predicted uninformative prompts before generation~\citep{zheng2025greso}. VIGOR estimates reward variance from initial rollouts and progressively allocates a fixed total budget to prompts with higher variance~\citep{jiang2026vigor}. Selection, filtering, and generation depth thus offer complementary ways to concentrate learning computation.

Table~\ref{tab:related-method} summarizes these distinctions. TAC and DUMP schedule predefined groups~\citep{yang2026tac,wang2025dump}, while FEAST learns capability groups without task labels. ADS favors clusters with higher observed success~\citep{xu2026ads}; FEAST preserves opportunities even when current utility is low. DAPO refills informative batches through extra generation~\citep{yu2025dapo}, whereas FEAST redistributes a fixed rollout budget across selected prompts.

%% file: tables/related_method.tex
\begin{table}[!b]
\caption{Comparison of data-sampling methods. Fairness and efficiency indicate whether these considerations are incorporated into the sampling design. Task-free means no supplied grouping labels. ClusterUCB uses target-task validation data. Extra generation includes refills and reference rollouts. $^{*}$SPaCe uses extra generation only for model-estimated difficulty. $^{\dagger}$BOTS requires extra generation only when reference scores are unavailable.}
\label{tab:related-method}
\centering
\setlength{\tabcolsep}{5pt}
\renewcommand{\arraystretch}{1.00}
\begin{tabular*}{\linewidth}{@{\extracolsep{\fill}}lcccc@{}}
\toprule
Method & \shortstack{Task-free} & \shortstack{Fairness} & \shortstack{Efficiency} & \shortstack{Extra\\generation} \\
\midrule
Skill-It~\citep{chen2023skillit} & -- & $\checkmark$ & -- & -- \\
ClusterUCB~\citep{wang2025clusterucb} & $\checkmark$ & $\checkmark$ & $\checkmark$ & -- \\
SPaCe~\citep{do2026space} & $\checkmark$ & $\checkmark$ & $\checkmark$ & $\checkmark^{*}$ \\
DAPO (dynamic sampling)~\citep{yu2025dapo} & $\checkmark$ & -- & $\checkmark$ & $\checkmark$ \\
GRESO~\citep{zheng2025greso} & $\checkmark$ & -- & $\checkmark$ & $\checkmark$ \\
TAC~\citep{yang2026tac} & -- & $\checkmark$ & -- & -- \\
ADS~\citep{xu2026ads} & $\checkmark$ & $\checkmark$ & $\checkmark$ & -- \\
DUMP~\citep{wang2025dump} & -- & $\checkmark$ & -- & -- \\
BOTS~\citep{shen2026bots} & $\checkmark$ & -- & $\checkmark$ & $\checkmark^{\dagger}$ \\
\midrule
\textbf{FEAST} & $\checkmark$ & $\checkmark$ & $\checkmark$ & -- \\
\bottomrule
\end{tabular*}
\end{table}

%% file: sec/3-qel.tex
\section{QEL: Why Allocation History Matters}
\label{sec:qel}
\label{sec:method/qel}

\noindent
\begin{minipage}[t]{0.43\linewidth}
\vspace{0pt}
\centering
\includegraphics[width=\linewidth]{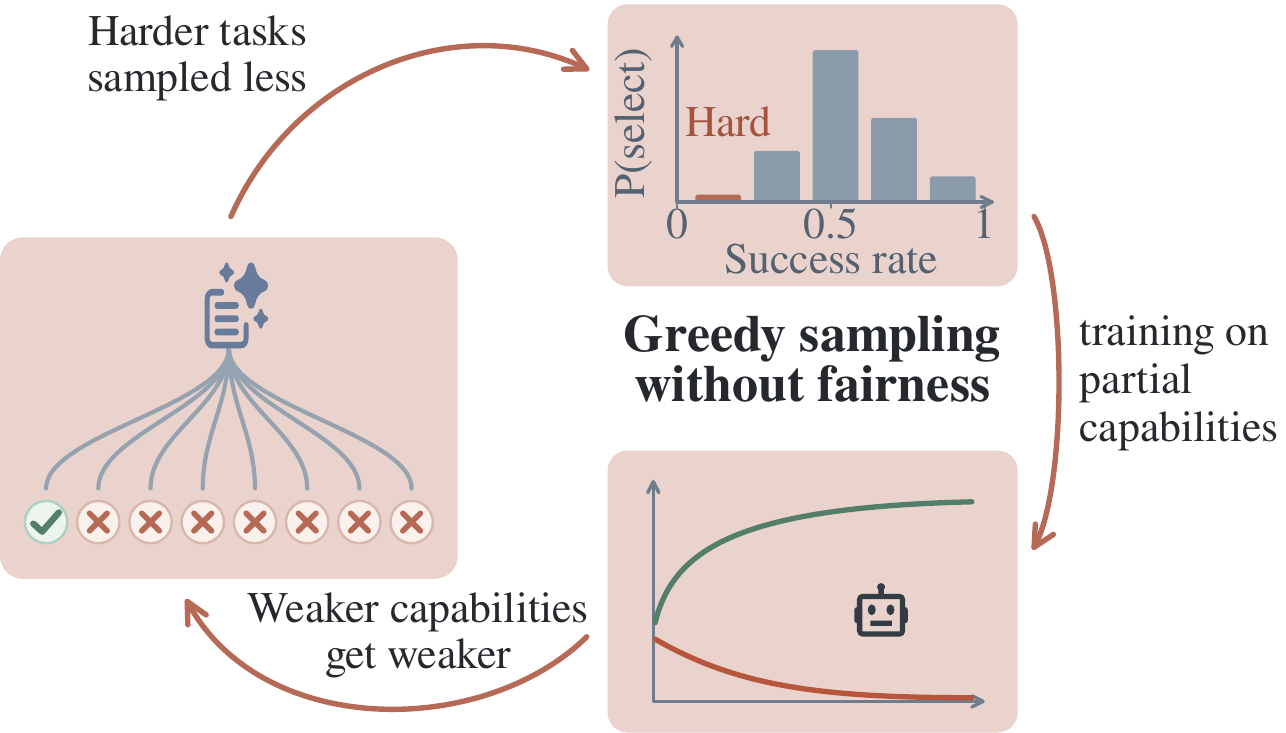}
\captionof{figure}{Greedy allocation can reinforce the neglect of weak capabilities.}
\label{fig:showing-qel}
\end{minipage}\hfill
\begin{minipage}[t]{\dimexpr0.57\linewidth-\columnsep\relax}
\vspace{0pt}
{\color{black}
When a policy learns multiple capabilities, does a fixed budget of training on capability $k$ always yield the same improvement? Let $V_k(\theta)$ denote the expected verifier reward of policy $\theta$ on a fixed evaluation set for $k$. Suppose schedule $\mathcal Q$ first accumulates $N$ training samples of $k$ (measured in prompt groups or rollouts) at step $T$. We define the matched-exposure learning efficiency as the average gain per sample:
{\small
\begin{equation}
\eta_k(N;\mathcal Q)=\frac{\mathbb E[V_k(\theta_T)]-V_k(\theta_0)}{N}.
\label{eq:qel-matched-exposure}
\end{equation}
}
Under a sparser schedule, the policy trains longer on other tasks before reaching $N$ samples of $k$. Matching $N$ separates allocation-history effects from differences in target-capability sample count.
}
\end{minipage}

\noindent\textbf{\rev{Equal exposure, unequal gains.}}
\rev{The value of a sample from capability $k$ depends on the current policy, which is shaped by all prior updates, including direct training on $k$ and cross-capability transfer. Consequently, the same $N$ samples can yield different gains depending on the allocation path. Theorem~\ref{thm:qel-decomposition} (Appendix~\ref{app:qel-theory}) formalizes this path dependence by decomposing $\eta_k(N;\mathcal Q)$ into cumulative direct learning and transfer along the trajectory.}

\noindent\textbf{\rev{When sparse allocation hurts.}}
\rev{Positive cross-capability transfer can sustain learning on $k$ between exposures. Under suppressive transfer, longer gaps can erode the policy's readiness to learn $k$ and reduce its matched-exposure efficiency (Proposition~\ref{prop:qel-staleness}). With persistent negative transfer, an effective quota below a sufficient threshold yields zero or negative expected net progress over initialization (Corollary~\ref{cor:qel-critical}).}

\noindent\textbf{\rev{From efficiency to fairness.}}
\rev{A capability that appears stagnant may be suffering from prior under-allocation, and further neglect can compound this deficit (Figure~\ref{fig:showing-qel}). Because learning efficiency depends on history, a scheduler should continue allocating budget to neglected capabilities so they can refresh their learning signals under the current policy. Learning-opportunity fairness requires that weak current signals never permanently exclude a capability, while allowing unequal quotas. Section~\ref{sec:method} builds FEAST on this principle.}

%% file: sec/4-feast.tex
\section{FEAST: Fair Across Clusters, Efficient Within}
\label{sec:method}

\begin{figure}[t]
\centering
\includegraphics[width=\linewidth]{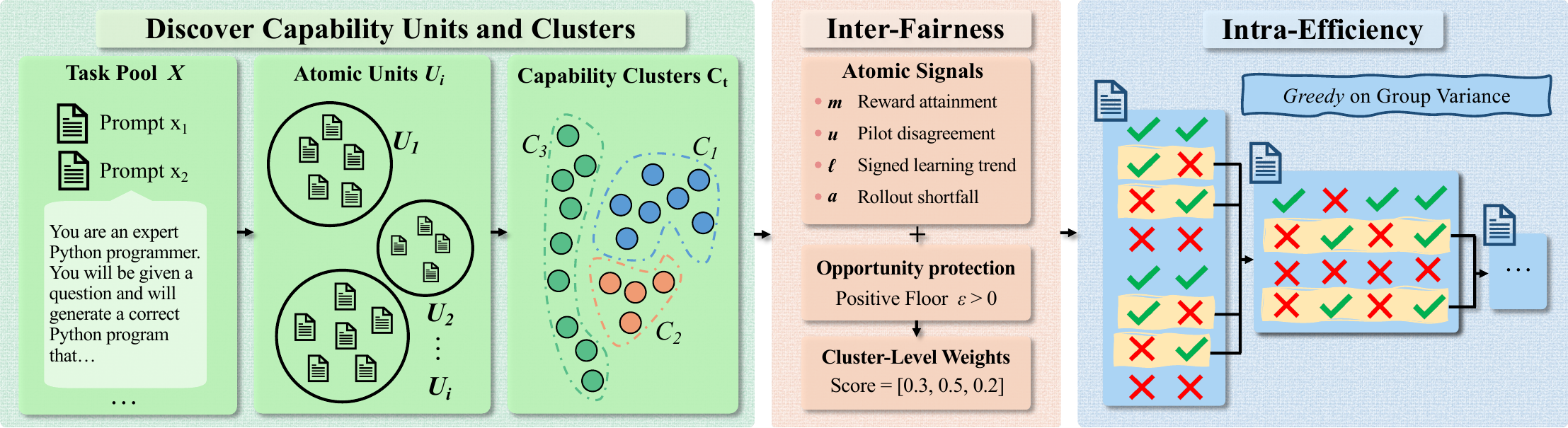}
\caption{\rev{\emph{Dynamic clustering} discovers capability clusters while retaining atomic-unit histories (Eq.~\ref{eq:feast-grouping}). \emph{Inter-cluster fairness} protects selection opportunities (Eqs.~\ref{eq:feast-priority}--\ref{eq:feast-inter}), and the \emph{reserved budgets mechanism} fixes each selected cluster's budget (Eq.~\ref{eq:feast-budget}). \emph{Intra-cluster efficiency} then progressively allocates rollouts within each budget (Eq.~\ref{eq:feast-intra}).} Pilot feedback and policy updates guide subsequent allocation and regrouping.}
\label{fig:pipeline-feast}
\end{figure}

Figure~\ref{fig:pipeline-feast} illustrates the FEAST framework. We consider a policy $\pi_\theta$, an unlabeled prompt pool $\mathcal X$, verifier feedback, and a fixed total budget of $B$ generated trajectories per update.

Let $\mathcal C_t$ be the current capability clusters and $\mathcal S_t\subseteq\mathcal C_t$ the $M$ clusters selected at update $t$. Each receives a disjoint budget $B_c=B/M$. Rollout counts $g_{x,t}$ for selected prompts $P_{c,t}$ satisfy
\begin{equation}
\sum_{x\in P_{c,t}}g_{x,t}=B_c,\qquad
\sum_{c\in\mathcal S_t}B_c=B.
\label{eq:feast-budget}
\end{equation}
Thus a cluster's opportunity is decided before its prompts compete for additional rollouts. \rev{We call this constraint the \emph{reserved budgets mechanism}. Once a cluster is selected, all of its $B_c$ rollouts, including pilots and later progressive rounds (Section~\ref{sec:method/intra-cluster}), are spent on its own prompts. High-variance prompts in other selected clusters therefore cannot take these rollouts, so a weak cluster keeps its full budget after selection.}

\subsection{Discovering Capabilities While Retaining History}
\label{sec:method/clustering}

FEAST separates evolving capability groups from persistent learning histories. An initial fine-grained partition $\mathcal U=\{U_i\}_{i=1}^{J}$ defines atomic units that retain their observations when regrouped. Neither atoms nor clusters need align with ground-truth task boundaries.

Periodically, FEAST represents each atomic unit using the current policy and groups units by both semantic similarity and learning dynamics according to
\begin{equation}
\mathcal C_t=\operatorname{Group}\!\left(\{[z_i(\theta_t),\bar\ell_{i,t}]\}_{i=1}^{J}\right),
\label{eq:feast-grouping}
\end{equation}
where $z_i(\theta_t)$ is its policy-dependent representation and $\bar\ell_{i,t}$ is a smoothed signed learning trend. Semantically related units can receive different priorities when their learning trends diverge. Representation extraction, clustering, and state updates are specified in Appendix~\ref{app:feast-details}.

\subsection{Fair Learning Opportunities Across Clusters}
\label{sec:method/inter-cluster}

For atomic unit $i$, let $m_{i,t}$ denote recent reward attainment, $u_{i,t}$ the mean squared difference between two pilot rewards, $\ell_{i,t}$ the exposure-normalized signed attainment trend, and $a_{i,t}$ the accumulated rollout shortfall from an equal per-update share, normalized and capped at 1. Its priority is
\begin{equation}
r_{i,t}=\underbrace{\epsilon+\lambda_a a_{i,t}}_{\text{opportunity protection}}
+\underbrace{u_{i,t}\,\Phi(m_{i,t})\bigl(1+\gamma|\ell_{i,t}|\bigr)}_{\text{learning-evidence surplus}},
\label{eq:feast-priority}
\end{equation}
where \rev{the exploration floor $\epsilon>0$ keeps every priority positive, $\lambda_a\geq0$ weights the rollout shortfall, $\gamma\geq0$ weights the learning trend,} and bounded $\Phi$ favors intermediate attainment with a modest preference for the less-mastered side. Absolute trends respond to improvement and regression. Attainment alone can conceal consistently successful and unsuccessful prompts with no within-prompt contrast, so FEAST measures within-prompt disagreement separately.

Cluster weights $w_{c,t}$ aggregate atomic priorities using counts of previously unsampled prompts and shrink toward $\epsilon$ as that capacity is exhausted (Appendix~\ref{app:feast-details}). Cluster sampling therefore uses temperature-scaled weights $\widetilde w_{c,t}=w_{c,t}^{1/\mathcal T}$, with $\mathcal T>0$.

The $M$ clusters are drawn without replacement. Let $\mathcal F_t$ denote the history before update $t$ and $\mathcal R_t$ the clusters remaining in the draw. The next cluster is sampled as
\begin{equation}
\Pr(c\text{ drawn next}\mid\mathcal F_t,\mathcal R_t)
=\frac{\widetilde w_{c,t}}{\sum_{c'\in\mathcal R_t}\widetilde w_{c',t}},
\qquad c\in\mathcal R_t.
\label{eq:feast-inter}
\end{equation}
Learning evidence changes selection frequency while retaining eligibility. \rev{Under the reserved budgets mechanism, prompt-level filtering cannot reassign selected budgets across clusters}, protecting each selected cluster's opportunity.

Uniformly positive, bounded weights give history-uniform positive selection probabilities for eligible current clusters, including after regrouping, and a geometric waiting-time bound (Proposition~\ref{prop:feast-service}). \rev{Together with the reserved budgets mechanism in \eqref{eq:feast-budget} and the rollout shortfall $a_{i,t}$, which also guides prompt allocation within each cluster (Appendix~\ref{app:feast-details}), this protects both cluster-level selection and atomic-unit exposure, as the experiments in Section~\ref{sec:experiments} and the service logs in Appendix~\ref{app:add-service} show.}

\subsection{Efficient Rollouts Within Each Opportunity}
\label{sec:method/intra-cluster}

Within each selected cluster, FEAST first samples a broad prompt set $P_c^{(0)}$ and gives each selected prompt a small, equal number of pilot rollouts. During subsequent rounds, it then retains prompts with the largest cumulative verifier-score variance:
\begin{equation}
P_c^{(r+1)}
=\operatorname{Top}_{\lceil\rho|P_c^{(r)}|\rceil}
\{v_{x,r}:x\in P_c^{(r)}\},
\qquad
\sum_r n_r|P_c^{(r)}|=B_c,
\label{eq:feast-intra}
\end{equation}
where $v_{x,r}$ is the empirical variance using all rollouts generated for prompt $x$ through round $r$, $\rho\in(0,1)$ is the retained fraction, and $n_r$ is the number of additional rollouts per surviving prompt. \rev{We call retaining a prompt for the next round \emph{promotion}.} The policy remains fixed throughout these rounds.

In reward-centered GRPO, identical sampled rewards yield zero task advantages; varying outcomes provide a relative signal~\citep{shao2024deepseekmath,yu2025dapo}, although regularization can remain active without task advantages~\citep{wang2026ragen2}. FEAST places variance-guided progressive allocation~\citep{jiang2026vigor} inside \rev{the reserved budgets of \eqref{eq:feast-budget}}: higher-variance prompts elsewhere cannot take a selected weak cluster's rollouts. Accordingly, every generated rollout, including both pilots and zero-signal groups, counts toward the total budget $B$.

FEAST estimates attainment and disagreement from common initial pilots, weighting prompts equally to avoid promotion bias. Additional rollouts refine training signals, and learning trends account for signal-bearing experience. Statistics remain attached to persistent units and guide regrouping and allocation together with the updated policy.

QEL shows that weak observed feedback can partly result from earlier under-allocation. A scheduler that uses this feedback to exclude a capability risks reinforcing the allocation pattern that produced it. FEAST preserves opportunities to reassess low-scoring clusters as the policy evolves. Within each selected cluster, prompts compete using current feedback, so preserving access to experience remains compatible with concentrating computation where useful learning signals are observed. \rev{Appendix~\ref{app:feast-walkthrough} walks through one update in the main configuration and gives the policy-loss aggregation.}

%% file: sec/5-experiments.tex
\section{Experiments}
\label{sec:experiments}

In this section, we first test QEL at matched task exposure, then evaluate FEAST's overall performance, computational cost, and individual components.

\subsection{QEL}
\label{sec:exp/qel}

\subsubsection{Experimental Setup}
\label{sec:exp/qel-setup}

We test the history dependence in Section~\ref{sec:qel} on Sokoban, Mastermind, and Countdown in TextArena~\citep{guertler2025textarena} with Qwen3-1.7B-Base. For each task, we compare uniform allocation ($1{:}1{:}1$) with a schedule that assigns it the smallest quota. Both schedules use the same task instances in the same order. Each update uses 32 prompt groups and eight rollouts per group. The two paths first reach $n=256$ task-specific prompt groups at updates 24 and 104, and $n=512$ at updates 48 and 208, respectively. As a result, the low-quota path trains longer to reach exactly the same task exposure, as required by the definition in Equation~\ref{eq:qel-matched-exposure}.

On Reasoning Gym~\citep{stojanovski2025reasoninggym}, we compare six fixed sampling ratios over Countdown, Spell backward, and Word sorting, aligning each task's held-out success by its cumulative prompt exposure. Each ratio trains for 400 updates. Evaluation uses a fixed panel of 256 held-out instances per task, shared across ratios, with one response per instance.

\subsubsection{Results and Analysis}
\label{sec:exp/qel-results}

\begin{figure}[!t]
\centering
\includegraphics[width=\textwidth]{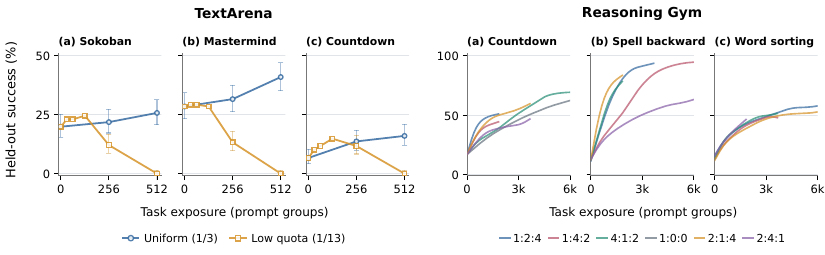}
\caption{QEL at matched task exposure. TextArena panels (left) compare uniform allocation with low-quota S:M:C mixtures $1{:}4{:}8$, $4{:}1{:}8$, and $8{:}4{:}1$, respectively, where S, M, and C denote Sokoban, Mastermind, and Countdown. Reasoning Gym panels (right) show six C:S:W sampling ratios for Countdown, Spell backward, and Word sorting. Curves are smoothed. Markers in the left panels show the observed scores with 95\% Wilson intervals.}
\label{fig:qel-matched-exposure}
\end{figure}

\noindent\textbf{Allocation history changes the return from equal task exposure.}
At $n=256$, low-quota paths in Figure~\ref{fig:qel-matched-exposure} trail uniform allocation by 9.77, 18.36, and 1.95 percentage points for Sokoban, Mastermind, and Countdown, respectively. At $n=512$, all three low-quota outcomes are zero, while uniform allocation improves each task over its initial policy. With common initialization and exposure denominators, these gaps also order empirical learning efficiencies, consistent with QEL's adverse direction and the need to protect opportunities.

\noindent\textbf{Sampling ratios change the average return per prompt.}
For Reasoning Gym, we measure held-out gain from each initialization per task-specific prompt (Equation~\ref{eq:qel-matched-exposure}). At $n=1829$, the $1{:}2{:}4$ and $1{:}4{:}2$ mixtures yield 18.58 versus 14.95 percentage points of gain per 1,000 prompts for Countdown, and 38.87 versus 18.15 for Spell backward (Figure~\ref{fig:qel-matched-exposure}). The larger Spell backward gap shows task-dependent sensitivity to allocation history even after normalizing exposure, consistent with the \rev{direct learning and signed cross-capability transfer} terms in Theorem~\ref{thm:qel-decomposition}.

Additional task sets, longer-horizon comparisons, and experimental details appear in Appendix~\ref{app:qel-additional-evidence}\rev{, and Appendix~\ref{app:add-qel-guru} repeats the matched-exposure comparison on GURU}.

\subsection{FEAST}
\label{sec:exp/feast}

\input{tables/feast_main}

\subsubsection{Experimental Setup}
\label{sec:exp/feast-setup}

We evaluate FEAST on GURU-18K~\citep{cheng2025guru} with Qwen3-1.7B-Base and Qwen3-8B-Base across mathematics, code, science, logic, simulation, and tabular reasoning. Baselines are uniform prompt sampling (Random), TAC~\citep{yang2026tac}, ADS~\citep{xu2026ads}, DUMP~\citep{wang2025dump}, and BOTS~\citep{shen2026bots}. All train one policy across the shared pool for 100 optimizer updates and share evaluation panels within each scale. TAC switches domains between single-domain updates; the others allow mixed-domain batches. TAC and DUMP use domain identities, whereas FEAST requires none. Avg.\ equally weights nine benchmark scores across the six domains. Appendix~\ref{app:experiment-details} gives per-seed results. We also evaluate Online Difficulty Filtering (ODF)~\citep{bae2026odf} on GURU-18K at 1.7B as an additional filtering baseline.

We also evaluate FEAST on SMC (Science, Math, and Code), which combines WildSci~\citep{liu2026wildsci}, MATH~\citep{hendrycks2021math}, and code from LeetCode, LiveCodeBench~\citep{jain2024livecodebench}, PrimeIntellect~\citep{mattern2025synthetic1}, and TACO~\citep{li2023taco}. Its 11,975 verifier-compatible training prompts span seven science categories, seven mathematics subjects, and one code category. Models trained separately on SMC are evaluated on 450 held-out prompts (30 per category, eight responses each), with terminal checkpoints at update 100 for 1.7B and 75 for 8B. Appendix~\ref{app:experiment-details} gives benchmark results, data construction, and protocols.

Across both suites, GRPO comparisons and ablations use 1,024 online rollouts per update, including pilots and zero-variance groups, with matched update counts per dataset and scale. BOTS's offline generation is accounted for separately; Section~\ref{sec:exp/efficiency} discusses refill-based comparisons.

\subsubsection{Results}
\label{sec:exp/feast-results}

\label{sec:exp/guru}

\noindent\textbf{Utility-driven sampling can reinforce neglect.}
Under matched online rollout budgets at 1.7B, ADS, BOTS, DUMP, and TAC trail Random in Avg.\ (Table~\ref{tab:guru18k-main}), consistent with Section~\ref{sec:qel}. On Logic, ADS, BOTS, and TAC trail by 32.79, 29.66, and 12.19 points; ADS also trails by 13.03 points on Tabular. A similar imbalance also remains evident at 8B: ADS leads these schedulers on Science, yet its Logic score remains 19.52 points below Random. At 1.7B, ADS leaves Logic near its initial score despite improving Science. This uneven progress is consistent with repeatedly favoring capabilities that already provide useful feedback. Periodic sampling of weak capabilities can refresh their learning signals as training on other domains changes the shared policy.

\noindent\textbf{Overall performance.}
As shown in Table~\ref{tab:guru18k-main}, FEAST achieves the highest average scores among the compared GRPO schedulers, reaching 36.18 at 1.7B and 51.69 at 8B and exceeding Random by 0.75 and 1.04 points, respectively. At 1.7B, FEAST exceeds Random by 4.58 points on Tabular but trails it by 2.57 points on Logic. At 8B, FEAST matches Random on Math and achieves the highest scores among these schedulers on Logic, Simulation, and Tabular, exceeding Random by 2.83, 4.56, and 1.82 points, respectively. These three domains have lower initial scores than Math and Code, and their gains are consistent with FEAST preserving learning opportunities across a broader range of capabilities. FEAST nevertheless scores 1.12 points below Random on Code at 8B, indicating that its aggregate gains still involve trade-offs across domains. Across the reported training seeds, FEAST also has a lower sample SD of Avg.\ than Random at both scales, suggesting more stable performance. \rev{Paired 95\% CIs of these differences lie above zero (Appendix~\ref{app:add-significance}).}

\noindent\textbf{Complementarity with DAPO.}
At 1.7B, DAPO+FEAST improves Avg.\ from 36.80 to 37.77, a gain of 0.97 points over DAPO, with improvements on Math, Code, Logic, and Tabular. At 8B, DAPO+FEAST achieves an Avg.\ score of 51.11 after 75 updates, compared with 50.59 for DAPO. This suggests that FEAST's allocation can complement DAPO's dynamic filtering. FEAST allocates the candidate rollouts, while DAPO selects which resulting groups contribute to the current policy update. In the combined method, feedback from candidate rollouts updates FEAST's learning histories even when their groups are excluded by filtering. These histories guide subsequent allocation before DAPO selects signal-bearing groups for optimization. Thus, FEAST can change the capability composition of the candidate pool while preserving DAPO's filtering criterion.

\input{tables/smc_results}

\label{sec:exp/supplement}
\label{sec:exp/smc}

\noindent\textbf{Results on SMC.}
FEAST exceeds Random by 3.33 points at 1.7B and 2.28 at 8B, outperforming ADS and DUMP at both scales (Table~\ref{tab:smc-indomain}). Acc.\ is average success (\%); TVA is mean gain over Random in the five least-improved categories. Negative TVA means that higher average performance need not improve every category. Among adaptive samplers, FEAST has the highest TVA at 1.7B and ties DUMP at 8B. Per-domain results are reported in Appendix Table~\ref{tab:smc-domains}.

\subsubsection{Computational Cost}
\label{sec:exp/feast-analysis}
\label{sec:exp/efficiency}

FEAST keeps the same per-update rollout budget as uniform sampling. Pilots and progressive rounds both draw from this budget, and all generated trajectories, including zero-variance groups, are counted. In the 1.7B GURU-18K run with seed 42, ten representation refreshes take 42.0 minutes in total, accounting for 7.1\% of the 9.90-hour training run.

BOTS~\citep{shen2026bots} additionally uses offline reference rollouts to construct its selection priors. DAPO~\citep{yu2025dapo} filters zero-variance groups, while the evaluated Online Difficulty Filtering (ODF)~\citep{bae2026odf} retains groups of intermediate empirical difficulty. Both filtering procedures generate extra candidates to fill optimization batches, so rejected groups still incur generation cost. ODF consumes 737.5M tokens over 50 updates but achieves an average score of only 31.92\%. Given its high generation cost and limited performance, we do not extend training to 100 updates. Appendix~\ref{app:experiment-details} provides full computational accounting\rev{, including end-to-end training GPU-hours,} and ODF settings.

\subsubsection{Ablation Study}
\label{sec:exp/ablation}

\input{tables/feast_ablation}
\textbf{\rev{Inter-cluster fairness and intra-cluster efficiency.}}
All ablations use \rev{GURU-18K with Qwen3-1.7B-Base, seed 42, 100 updates, and a fixed rollout budget}. \rev{FEAST w/o inter-cluster fairness samples prompts globally but} retains progressive allocation (Table~\ref{tab:guru18k-ablation}). Its lower score supports reserving capability-level budgets before prompts compete for additional rollouts. \rev{FEAST w/o intra-cluster efficiency, which fixes eight rollouts per prompt,} produces a larger decline, supporting the use of current reward variation to distribute computation within each reserved budget. These losses therefore support combining capability-level protection with selective computation within each budget. \rev{The remaining rows remove one mechanism at a time. Every removal lowers Avg., indicating that each tested mechanism contributes. FEAST w/o reserved budgets mechanism, in which high-variance prompts in one selected cluster can take another's rollouts, gives the largest drop. Appendix~\ref{app:experiment-details} defines each variant and varies the number of atomic units.}

%% file: tables/feast_main.tex
\begin{table}[!t]
\centering
\normalsize
\setlength{\tabcolsep}{2pt}
\newcommand{\feastscore}[2]{\mbox{#1{\fontsize{8}{9}\selectfont$\pm$#2}}}
\renewcommand{\arraystretch}{1.12}
\caption{GURU-18K results (\%). Cells show mean $\pm$ sample SD across multiple training seeds (1.7B: seeds 42/43/44; 8B: seeds 42/43), each evaluated with avg@$k$.}
\label{tab:guru18k-main}
\begin{tabular*}{\linewidth}{@{\extracolsep{\fill}}l*{7}{c}@{}}
\toprule
Method & Math & Code & Science & Logic & Sim. & Tabular & Avg. \\
\midrule
\multicolumn{8}{c}{\textit{Qwen3-1.7B-Base}} \\
Base & \mbox{51.40} & \mbox{49.29} & \mbox{13.66} & \mbox{1.46} & \mbox{2.88} & \mbox{11.18} & \mbox{22.67} \\
\cmidrule(lr){1-8}
Random & \feastscore{61.00}{0.87} & \feastscore{61.85}{0.70} & \feastscore{26.85}{0.99} & \feastscore{\textcolor{resultbest}{\textbf{34.24}}}{0.66} & \feastscore{\textcolor{resultworst}{\textbf{4.25}}}{0.33} & \feastscore{20.99}{0.76} & \feastscore{35.43}{0.18} \\
ADS & \feastscore{\textcolor{resultworst}{\textbf{58.27}}}{2.02} & \feastscore{\textcolor{resultworst}{\textbf{60.01}}}{1.26} & \feastscore{\textcolor{resultbest}{\textbf{28.22}}}{0.38} & \feastscore{\textcolor{resultworst}{\textbf{1.45}}}{1.14} & \feastscore{\textcolor{resultworst}{\textbf{4.25}}}{0.50} & \feastscore{\textcolor{resultworst}{\textbf{7.96}}}{0.67} & \feastscore{\textcolor{resultworst}{\textbf{28.48}}}{0.32} \\
BOTS & \feastscore{62.80}{1.83} & \feastscore{\textcolor{resultbest}{\textbf{61.88}}}{0.53} & \feastscore{27.52}{0.82} & \feastscore{4.58}{1.70} & \feastscore{4.92}{0.44} & \feastscore{24.16}{1.38} & \feastscore{33.27}{0.30} \\
DUMP & \feastscore{60.60}{2.91} & \feastscore{61.71}{0.62} & \feastscore{26.72}{0.26} & \feastscore{25.31}{14.67} & \feastscore{\textcolor{resultbest}{\textbf{5.38}}}{0.76} & \feastscore{22.72}{2.24} & \feastscore{34.84}{0.95} \\
TAC & \feastscore{\textcolor{resultbest}{\textbf{64.13}}}{0.42} & \feastscore{60.74}{0.45} & \feastscore{\textcolor{resultworst}{\textbf{25.46}}}{0.90} & \feastscore{22.05}{7.20} & \feastscore{4.88}{0.78} & \feastscore{\textcolor{resultbest}{\textbf{25.78}}}{0.81} & \feastscore{35.00}{1.16} \\
\textbf{FEAST} & \feastscore{60.40}{2.25} & \feastscore{61.81}{0.49} & \feastscore{27.00}{1.38} & \feastscore{31.67}{1.97} & \feastscore{4.79}{0.36} & \feastscore{25.57}{1.60} & \feastscore{\textcolor{resultbest}{\textbf{36.18}}}{0.13} \\
\cmidrule(lr){1-8}
DAPO & \feastscore{60.20}{2.31} & \feastscore{62.30}{0.79} & \feastscore{28.45}{1.44} & \feastscore{33.57}{0.10} & \feastscore{4.25}{0.33} & \feastscore{25.85}{0.41} & \feastscore{36.80}{0.53} \\
DAPO+FEAST & \feastscore{61.33}{1.53} & \feastscore{63.28}{0.75} & \feastscore{28.14}{0.41} & \feastscore{34.50}{0.77} & \feastscore{4.00}{0.33} & \feastscore{28.64}{1.03} & \feastscore{37.77}{0.42} \\
\midrule
\multicolumn{8}{c}{\textit{Qwen3-8B-Base}} \\
Base & \mbox{73.00} & \mbox{77.08} & \mbox{34.03} & \mbox{2.77} & \mbox{8.50} & \mbox{32.02} & \mbox{41.17} \\
\cmidrule(lr){1-8}
Random & \feastscore{\textcolor{resultbest}{\textbf{80.40}}}{2.26} & \feastscore{80.81}{0.32} & \feastscore{40.36}{2.54} & \feastscore{35.24}{0.84} & \feastscore{6.38}{1.24} & \feastscore{45.73}{0.15} & \feastscore{50.65}{1.01} \\
ADS & \feastscore{\textcolor{resultworst}{\textbf{78.30}}}{0.71} & \feastscore{80.35}{0.62} & \feastscore{\textcolor{resultbest}{\textbf{41.47}}}{0.31} & \feastscore{\textcolor{resultworst}{\textbf{15.72}}}{1.32} & \feastscore{5.94}{0.44} & \feastscore{\textcolor{resultworst}{\textbf{41.48}}}{4.43} & \feastscore{\textcolor{resultworst}{\textbf{47.40}}}{1.19} \\
BOTS & \feastscore{80.00}{1.70} & \feastscore{80.55}{0.68} & \feastscore{40.71}{0.73} & \feastscore{28.53}{0.59} & \feastscore{7.56}{2.21} & \feastscore{46.95}{0.26} & \feastscore{50.28}{0.05} \\
DUMP & \feastscore{79.30}{0.71} & \feastscore{80.13}{0.70} & \feastscore{40.20}{1.74} & \feastscore{37.71}{0.76} & \feastscore{\textcolor{resultworst}{\textbf{5.88}}}{0.88} & \feastscore{46.21}{0.96} & \feastscore{50.66}{0.54} \\
TAC & \feastscore{78.90}{0.99} & \feastscore{\textcolor{resultbest}{\textbf{80.93}}}{0.21} & \feastscore{\textcolor{resultworst}{\textbf{40.03}}}{0.70} & \feastscore{37.24}{1.79} & \feastscore{8.00}{3.36} & \feastscore{45.92}{0.92} & \feastscore{50.88}{0.05} \\
\textbf{FEAST} & \feastscore{\textcolor{resultbest}{\textbf{80.40}}}{0.28} & \feastscore{\textcolor{resultworst}{\textbf{79.69}}}{0.10} & \feastscore{40.67}{1.26} & \feastscore{\textcolor{resultbest}{\textbf{38.07}}}{0.66} & \feastscore{\textcolor{resultbest}{\textbf{10.94}}}{3.27} & \feastscore{\textcolor{resultbest}{\textbf{47.55}}}{0.14} & \feastscore{\textcolor{resultbest}{\textbf{51.69}}}{0.08} \\
\cmidrule(lr){1-8}
DAPO & \feastscore{78.00}{2.26} & \feastscore{80.80}{0.21} & \feastscore{39.47}{0.70} & \feastscore{38.54}{0.20} & \feastscore{6.81}{0.97} & \feastscore{45.71}{1.64} & \feastscore{50.59}{0.64} \\
DAPO+FEAST & \feastscore{78.50}{0.99} & \feastscore{81.26}{0.63} & \feastscore{40.75}{0.97} & \feastscore{38.51}{0.36} & \feastscore{6.25}{0.53} & \feastscore{46.34}{0.28} & \feastscore{51.11}{0.23} \\
\bottomrule
\end{tabular*}
\end{table}

%% file: tables/smc_results.tex
\Needspace{12\baselineskip}
\begin{wraptable}{r}{0.46\textwidth}
\vspace{-7mm}
\centering
\small
\setlength{\tabcolsep}{4pt}
\caption{SMC results (avg@8).}
\label{tab:smc-indomain}
\begin{tabular*}{\linewidth}{@{\extracolsep{\fill}}lrrrr@{}}
\toprule
& \multicolumn{2}{c}{1.7B} & \multicolumn{2}{c}{8B} \\
\cmidrule(lr){2-3}\cmidrule(lr){4-5}
Method & Acc. & TVA & Acc. & TVA \\
\midrule
Random & 61.42 & 0.00 & 78.42 & 0.00 \\
ADS & 58.86 & $-8.58$ & 77.19 & $-5.42$ \\
DUMP & 63.47 & $-3.67$ & 78.75 & $-2.25$ \\
\textbf{FEAST} & 64.75 & $-2.50$ & 80.69 & $-2.25$ \\
\bottomrule
\end{tabular*}
\end{wraptable}

%% file: tables/feast_ablation.tex
\begin{wraptable}{r}{0.47\textwidth}
\vspace{-\dimexpr\intextsep+\abovecaptionskip\relax}
\centering
\caption{FEAST ablations. \rev{Each w/o row removes the named mechanism of Section~\ref{sec:method} from FEAST.} $\Delta$ is relative to the full FEAST configuration.}
\label{tab:guru18k-ablation}
\small
\setlength{\tabcolsep}{3pt}
\begin{tabular*}{\linewidth}{@{\extracolsep{\fill}}lrr@{}}
\toprule
Method & Avg. & $\Delta$ \\
\midrule
\textbf{FEAST} & 36.32 & $N/A$ \\
\rev{w/o inter-cluster fairness} & 35.66 & $-0.66$ \\
\rev{w/o intra-cluster efficiency} & 34.66 & $-1.67$ \\
\rev{w/o reserved budgets mechanism} & \rev{32.56} & \rev{$-3.76$} \\
\rev{w/o capability clusters} & \rev{34.01} & \rev{$-2.31$} \\
\rev{w/o dynamic clustering} & \rev{35.65} & \rev{$-0.67$} \\
\rev{w/o exploration floor ($\epsilon=0$)} & \rev{35.70} & \rev{$-0.62$} \\
\rev{w/o rollout shortfall ($\lambda_a=0$)} & \rev{35.94} & \rev{$-0.38$} \\
\rev{w/o learning trend ($\gamma=0$)} & \rev{35.85} & \rev{$-0.47$} \\
\bottomrule
\end{tabular*}
\end{wraptable}

%% file: sec/6-conclusion.tex
\section{Conclusion}
\label{sec:conclusion}

We presented FEAST, a task-free RL framework that combines fairness across capability clusters with efficiency within each cluster. Our QEL analysis shows that under suppressive conditions, allocation history can reduce learning gains even at matched capability-specific exposure. Experiments on GURU-18K and SMC with 1.7B and 8B models show improved average performance over other sampling methods under matched rollout budgets, with ablations supporting both allocation levels. These findings support learning-opportunity protection in on-policy self-evolution. \rev{Matched-exposure experiments on GURU and single-mechanism ablations further confirm that allocation history shapes learning efficiency and that each component contributes to overall performance.}

\section*{AI Use Statement}
We used generative AI tools to improve the language and readability of the manuscript and to assist with code implementation and debugging. The authors take full responsibility for the final text, code, and reported results, including all AI-assisted contributions.

%% file: sec/appendix.tex
\begingroup
\setlength{\LTcapwidth}{\textwidth}
\section{Appendix}

\begingroup
\flushbottom
\input{sec/appendix-qel}
\endgroup

\input{sec/appendix-feast}

\subsection{Experimental Protocol and Additional Results}
\label{app:experiment-details}

\paragraph{GURU-18K construction.}
We construct a fixed pool with 3,000 examples from each of mathematics, code, science, logic, simulation, and tabular reasoning, following the per-domain sampling rule of GURU~\citep{cheng2025guru}. Five domains come from the GURU-92K release at revision \texttt{0f7f6e8fc2cd36d46a9629310c9a4f9d8c1ffa3d}. The science domain is replaced by a pinned NVIDIA OpenScienceReasoning-2~\citep{nvidia2025opensciencereasoning2} subset (revision \texttt{174b02c9cdf231f220765b2a1d5ece4550921894}). Input-length filtering retains 17,067 of the 18,000 prompts. The same retained pool is used for each compared scheduler, and evaluation domain labels are not supplied to FEAST.

\paragraph{Training and comparison scope.}
The GRPO comparisons use Qwen3-1.7B-Base and Qwen3-8B-Base, training seeds 42/43/44 at 1.7B and seeds 42/43 at 8B, and the checkpoint after 100 optimizer updates at both scales. DAPO and DAPO+FEAST use 100 updates at 1.7B and 75 at 8B, while ODF uses 50 updates at 1.7B. Each GRPO update samples 128 prompts and generates exactly 1,024 online trajectories, including pilots and zero-variance groups. Random, TAC, ADS, DUMP, and BOTS use eight rollouts per prompt, while FEAST and FEAST w/o inter-cluster fairness redistribute the same budget through progressive rounds. We use GRPO with learning rate $5\times10^{-7}$, weight decay 0.1, PPO mini-batches of eight prompt groups, clipping bounds 0.2/0.2, and disabled reward KL and actor KL loss. Training prompt and response limits are both 2,048 tokens, with temperature 1.0 and top-$p$ 1.0. Training and evaluation share the domain-specific verifier semantics. Code is checked with the pinned Reasoning360 \texttt{coder1} path and a local SandboxFusion-compatible service, with a 30-second program timeout. Science uses strict boxed multiple-choice scoring.

TAC and DUMP receive the six domain identities. TAC selects one domain for each update and samples all 128 prompts from that domain, alternating domains over training. The other schedulers allow batches to mix domains without requiring per-update coverage of all six domains. ADS uses 62 semantic clusters obtained by merging small clusters after a 64-cluster initialization. FEAST uses the persistent atomic units and current-policy regrouping described in Appendix~\ref{app:feast-details}, with raw pilot rewards, squared pilot-reward differences, rollout shortfalls, log-responsibility GMM repair, absolute progress in its priority, and progressive increments $[2,4,16]$. FEAST w/o inter-cluster fairness removes the capability structure and samples 128 prompts globally without replacement before applying the progressive allocation. FEAST w/o intra-cluster efficiency retains FEAST's structure and allocation across clusters but fixes every response group at eight. \rev{FEAST w/o reserved budgets mechanism keeps FEAST's cluster selection and pilot rollouts but runs the later variance-based promotion over all selected prompts together, so high-variance prompts in one cluster can divert rollouts from other selected clusters. FEAST w/o capability clusters applies inter-cluster fairness directly to atomic units and selects four units with 32 prompts each. FEAST w/o dynamic clustering keeps the initial clustering throughout training, and FEAST w/o exploration floor, FEAST w/o rollout shortfall, and FEAST w/o learning trend set $\epsilon$, $\lambda_a$, and $\gamma$ in Equation~\ref{eq:feast-priority} to zero, respectively.}

DAPO+FEAST combines FEAST's grouping and atomic priorities with eight rollouts per prompt, refilling until 128 signal-bearing prompt groups are retained. Selected clusters and atomic quotas remain fixed across refill batches. Attainment and productivity use the first two raw rewards of every candidate prompt, including rejected groups. The trend denominator counts trajectories retained for optimization. The shortfall increment equals the total generated trajectories per update divided by $J$, including refill, while its normalization remains $10\times1024/J$.

BOTS uses the official interpolation-based Bayesian selector with target success 0.5, forgetting coefficient 0.1, and implicit-evidence weight 0.1. Its prior features pair Qwen3-1.7B (the post-trained model) with Qwen2.5-7B-Instruct. We generate 16 reference responses for all 18,000 prompts with Qwen3-1.7B and for the 3,000 substituted science prompts with Qwen2.5-7B-Instruct, using a 4,096-token response cap. The other 15,000 prompts reuse published 7B pass-rate features. Newly generated reference evaluations therefore comprise 336,000 responses. The completed feature artifact is shared across the two policy scales.

\input{tables/guru18k_evaluation}
\paragraph{Evaluation panels and metrics.}
Evaluation panels are taken from GURU~\citep{cheng2025guru}. The nine primary benchmarks use input cap 4,096 and evaluation seed 42. MATH-500~\citep{lightman2023verify}, HumanEval~\citep{chen2021evaluating}, MBPP~\citep{austin2021program}, GPQA-Diamond~\citep{rein2023gpqa}, and SuperGPQA~\citep{map2025supergpqa} use temperature 0.7 and top-$p$ 0.8, while the other four benchmarks use temperature 1.0 and top-$p$ 0.7. Scores average all sampled responses per question. Table~\ref{tab:guru18k-evaluation} specifies the panel sizes and response caps for the primary and supplementary evaluations.

Zebra~\citep{cheng2025guru} uses official cell-level partial credit. FinQA~\citep{chen2021finqa} retains eight duplicate question--answer pairs and their original weights, giving 1,093 distinct identities among 1,101 records. CodeIO~\citep{li2025codeio} retains the official NaN reference at record 103, which receives zero.

\paragraph{Benchmark selection and supplementary evaluations.}
The main table reports MATH-500 for mathematics and the code benchmarks HumanEval and MBPP. GPQA-Diamond and SuperGPQA cover science, Zebra covers logic, CodeIO covers simulation, and FinQA and \mbox{MultiHierTT~\citep{zhao2022multihiertt}} cover tabular reasoning. These nine benchmarks span all six training domains and have complete evaluations for the reported checkpoints, allowing each method and seed to use the same aggregate.

The supplementary math panels use input cap 4,096, response cap 8,192, temperature 0.7, top-$p$ 0.8, and evaluation seed 42. ARC400~\citep{chollet2019measure} is the 400-question ARC-AGI-1 panel, scored by exact-grid accuracy. Its input cap is 24,576 to retain all prompts within the 32,768-token context. Ordering100 uses input cap 4,096 and strict ordering accuracy. Both use temperature 1.0, top-$p$ 0.7, and evaluation seed 42. For Ordering100, we report both the official score and a case-normalized score. The official scorer lowercases predictions but not reference answers. The case-normalized score applies the same normalization to the references while reusing the original responses.

\paragraph{Aggregation and uncertainty.}
Math, Logic, and Simulation report MATH-500, Zebra, and CodeIO, respectively. Code averages HumanEval and MBPP, Science averages GPQA-Diamond and SuperGPQA, and Tabular averages FinQA and MultiHierTT. Avg.\ equally weights the nine benchmarks, equivalent to domain weights $1{:}2{:}2{:}1{:}1{:}2$. All aggregates use unrounded scores. Green and red mark the highest and lowest displayed means among the GRPO schedulers, including ties.

Table~\ref{tab:guru18k-main} reports mean $\pm$ sample standard deviation across training seeds 42/43/44 for 1.7B methods with three completed training seeds and across seeds 42/43 for all trained 8B methods, using denominator $n-1$. Domain and overall scores are computed separately for each seed before calculating their means and SDs. Tables~\ref{tab:guru18k-seeds} and~\ref{tab:guru18k-seeds-8b} list each checkpoint's nine primary benchmark scores and AMC for 1.7B and 8B, respectively. In the detailed tables, MATH, H.Eval, GPQA, S.GPQA, and M.Hier abbreviate MATH-500, HumanEval, GPQA-Diamond, SuperGPQA, and MultiHierTT, respectively. Base rows evaluate the original, untrained checkpoints at update zero and omit error bars. At 1.7B, ODF uses training seed 42.

Ablations use seed 42 at 1.7B with a matched full-method reference.

\input{tables/feast_benchmarks}
\input{tables/feast_benchmarks_8b}

\input{tables/feast_odf}
\paragraph{ODF results.}
\label{app:odf}
We evaluate the balanced filtering variant of ODF~\citep{bae2026odf} on GURU-18K with Qwen3-1.7B-Base and training seed 42, using the same optimizer settings and evaluation panels as DAPO. Each candidate prompt receives eight responses, and we retain the complete group when its empirical success rate satisfies $0.3<\hat p<0.7$, equivalent to three through five correct responses. We use DAPO's synchronous refill backend to collect 128 accepted groups per update, with a maximum of 64 candidate batches. We evaluate the update-50 checkpoint and count all generated responses, including rejected and excess groups, toward the reported 841,728 rollouts and 737.5M tokens (Table~\ref{tab:guru18k-odf}).

\paragraph{Online token accounting.}
Table~\ref{tab:guru18k-computation} reports computation for training seed 42 at the reported checkpoints. Online totals include non-padding prompt and response tokens for every generated trajectory, including pilots, progressive rounds, zero-variance groups, and refill attempts. Prompt tokens are counted once per rollout, without prefix-cache discounts. At 1.7B, the totals are 91.3M (Random), 107.6M (TAC), 74.5M (ADS), 94.3M (DUMP), 92.3M (BOTS), 101.4M (FEAST) and 357.8M (DAPO). DAPO uses clipping 0.2/0.28 with clip-$c$ 10, no KL term, and at most 64 refill batches per update. It retains 128 groups of eight responses per update, generating 425,984 trajectories in total. At 8B, DAPO+FEAST generates 206,848 trajectories and consumes 201.0M tokens over 75 updates. Totals follow the retained training history.

The shared BOTS preparation artifact contains 336,000 newly generated reference responses, totaling 1,149.8M measured tokens, and is used at both policy scales.

\paragraph{\rev{End-to-end training cost.}}
\label{app:add-cost}
\rev{Table~\ref{tab:add-cost} reports end-to-end training GPU-hours for Qwen3-1.7B-Base over 100 updates, each run using one A100-80GB GPU on the same machine. All six methods keep their own prompt selection and response allocation while sharing the same inference and training configurations and generating the same 1,024 responses per update. FEAST's own overhead, representation extraction and regrouping, takes about 7\% of its training time (Section~\ref{sec:exp/efficiency}) and has little effect on end-to-end cost. The generated tokens of the selected prompts and post-generation filtering affect end-to-end cost much more. In these timed runs, ADS processes 70.86M prompt-plus-response tokens over the same 102,400 generated responses, compared with Random's 96.37M. After verification, ADS and DUMP discard groups whose responses all have zero accuracy, FEAST discards groups with zero reward variance, and Random, TAC, and BOTS keep all groups. Filtering reduces the tokens passed to log-probability computation and actor updates without changing the generation budget.}

\begingroup
\input{tables/feast_computation}
\begin{table}[!htbp]
\color{black}
\centering
\small
\caption{Estimated end-to-end training GPU-hours for Qwen3-1.7B-Base over 100 updates, with 128 prompts and 1,024 generated responses per update before filtering. Times include representation refresh and exclude evaluation and BOTS offline preparation.}
\label{tab:add-cost}
\begin{tabular}{lcccccc}
\toprule
 & Random & TAC & ADS & DUMP & BOTS & FEAST \\
\midrule
GPU-hours & 6.82 & 8.59 & 3.51 & 3.99 & 5.91 & 6.07 \\
\bottomrule
\end{tabular}
\end{table}

\Needspace{6\baselineskip}
\paragraph{Number of atomic units.}
\label{app:atomic-units}
\input{tables/feast_atomic_units}
Table~\ref{tab:guru18k-atomic-units} varies the number of persistent atomic units. Coarser units pool more prompts into each learning history, and regrouping can only move entire units. The lower score at $J=32$ is consistent with this loss of resolution. With a fixed rollout budget, increasing $J$ reduces the average exposure per unit. Learning trends may then rely on fewer observations, limiting the benefit of finer partitions. The default $J=64$ achieves performance comparable to $J=128$ with fewer learning histories to maintain and fewer atomic units to regroup.

\Needspace{6\baselineskip}
\paragraph{SMC protocol.}
The SMC training pool contains 11,975 prompts across 15 categories. Evaluation uses 30 questions per category, eight responses per question, temperature 0.7, top-$p$ 0.8, and a 14,336-token response cap. All methods use training seed 42, with final checkpoints at update 100 for 1.7B and update 75 for 8B. Each compared GRPO scheduler generates 1,024 online rollouts per update, including pilots and zero-variance groups, giving 102,400 rollouts at 1.7B and 76,800 at 8B.

\input{tables/smc_domains}
\input{figures/sec_greedy_starvation}

For category $k$, let $\delta_{m,k}=r_{m,k}-r_{\mathrm{Random},k}$ denote the success-rate difference at the same model scale and checkpoint. Sorting these 15 differences in ascending order, Tail Value-Added is
\begin{equation}
\label{eq:tva}
\mathrm{TVA}_{1/3}(m)=\frac{100}{5}\sum_{i=1}^{5}\delta_{m,(i)}.
\end{equation}
Thus the tail consists of the five least-improved categories.

Self-Evolving Curriculum (SEC)~\citep{chen2026sec} updates category scores using temporal-difference learning with mean absolute advantage as feedback, then samples categories from a softmax distribution over these scores. Figure~\ref{fig:sec-greedy-starvation} shows SEC-Greedy, a diagnostic variant on SMC that replaces this stochastic selection with deterministic selection of the four highest-scoring categories after a 15-step balanced warmup. Its left panel reports held-out success, and its right panel counts class dispatches after warmup. Code receives no further dispatches, and Algebra and Prealgebra receive three each. These allocation patterns illustrate how greedy selection can deprive weak capabilities of fresh feedback, motivating FEAST's explicit protection of learning opportunities.

\endgroup
\input{sec/appendix-domain-trajectories}

\Needspace{6\baselineskip}
\subsection{Additional QEL Evidence}
\label{app:qel-additional-evidence}

\begin{figure}[!tbp]
\centering
\includegraphics[width=0.89\linewidth,trim=0 7bp 0 0,clip]{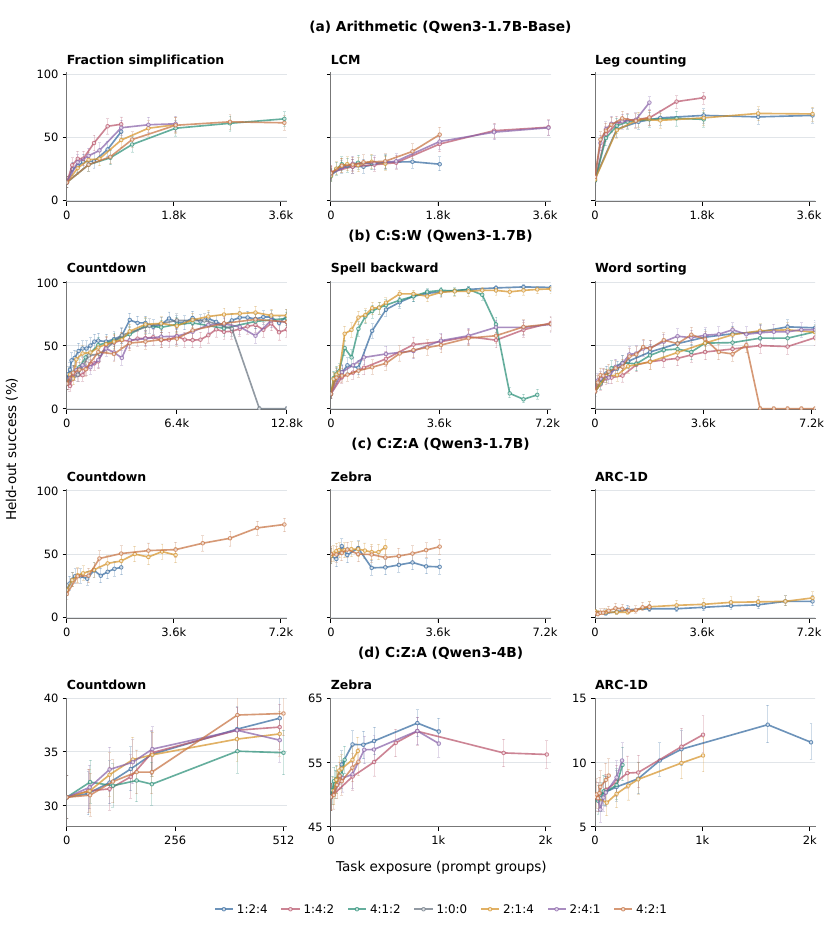}
\caption{Additional QEL results at matched exposure; ratios follow each row's task order. Markers show unsmoothed observations: avg@1 in (a)--(c), avg@8 in (d). Error bars show pointwise 95\% response-level Wilson intervals on fixed evaluation panels.}
\label{fig:qel-appendix-grid}
\label{fig:qel-arithmetic-context}
\label{fig:qel-csw-long-horizon}
\label{fig:qel-rg-eval-success-by-exposure}
\label{fig:qel-cza-4b}
\end{figure}

\paragraph{Main-figure protocol.}
For the TextArena comparisons in Figure~\ref{fig:qel-matched-exposure}, both schedules use the same task instances in the same order for each task. We compute 95\% Wilson intervals over 256 held-out instances per task. The Reasoning Gym panels in that figure use six completed 400-update runs, with zero-quota task--mixture pairs omitted. Their display smoothing applies a weighted three-point average $(0.2,0.6,0.2)$ followed by a Gaussian filter with bandwidth 350 prompt groups. Odd reflection preserves the observed endpoints; the numerical comparisons use the unsmoothed observations.

The additional Reasoning Gym~\citep{stojanovski2025reasoninggym} comparisons use 256 fixed held-out instances per task. The 1.7B experiments use one response per instance; the 4B experiment averages eight responses (avg@8). Each experiment retains its own panel, shared across ratios. Markers show observed checkpoints, connected without smoothing; numerical comparisons use exactly attained exposures.

\paragraph{Task composition matters at fixed target-task quota.}
Figure~\ref{fig:qel-appendix-grid}(a) compares all six permutations of $1{:}2{:}4$ over fraction simplification, least common multiple (LCM), and leg counting. These runs use Qwen3-1.7B-Base for 200 updates, with 32 prompt groups and eight rollouts per group, learning rate $10^{-5}$, LoRA rank and scaling 32, and entropy coefficient 0.001. Evaluation uses temperature 0.6, top-$p$ 0.95, and top-$k$ 20.

At update 200, fraction simplification reaches 54.30\% under $1{:}2{:}4$ and 60.16\% under $1{:}4{:}2$. Both paths allocate it quota $1/7$ and exactly 914 prompt groups. Leg counting reaches 81.25\% under $1{:}4{:}2$ and 64.06\% under $4{:}1{:}2$, both with quota $2/7$ and exactly 1,828 groups. Within each of these two pairs, the ordered target-task prompts are identical. Changing the companion-task allocation changes the policy path and the return from the same target-task sequence, consistent with the signed transfer and path-dependent alignment terms in Theorem~\ref{thm:qel-decomposition}.

\paragraph{Allocation-history gaps persist during extended training.}
Figure~\ref{fig:qel-appendix-grid}(b) uses a separate collection of Qwen3-1.7B runs on Countdown, Spell backward, and Word sorting (C:S:W). Training starts afresh with the first 400 updates' data preserved from the shorter protocol. The $1{:}2{:}4$ and $1{:}4{:}2$ runs reach 2,800 updates, the other mixed ratios reach 1,600, and the Countdown-only control reaches 400. The batch, rollout, learning-rate, LoRA, entropy, and evaluation settings match the preceding experiment. After a task/difficulty pool is exhausted, its fixed prompt permutation cycles. The longer-horizon runs use a separate frozen evaluation panel.

At 6,400 Spell backward groups and update 1,400, $2{:}1{:}4$ yields 93.75\%, whereas $4{:}1{:}2$ yields 7.42\%. Both assign Spell backward quota $1/7$; the latter path had reached 94.53\% at update 1,000 before declining. Conversely, at 12,800 Countdown groups, the single-task control scores 0.00\% and $1{:}2{:}4$ scores 73.44\%.

\begingroup
\setlength{\intextsep}{6pt}
\input{figures/ratio_allocation_tradeoff}
\paragraph{QEL comparisons on an additional task set.}
Figure~\ref{fig:qel-appendix-grid}(c) compares three allocation ratios on Countdown, Zebra, and ARC-1D (C:Z:A) over 400 training updates.

\paragraph{QEL comparisons with a 4B model.}
Figure~\ref{fig:qel-appendix-grid}(d) adds six completed Qwen3-4B runs on C:Z:A, covering all permutations of $1{:}2{:}4$. Training uses 32 prompt groups per update, eight rollouts per group, learning rate $10^{-5}$, LoRA rank and scaling 32, and entropy coefficient 0.001. Runs stop after 28, 55, or 110 updates, when Countdown exposure first reaches at least 500 groups; the attained counts are 503 or 512. Evaluation averages eight responses per instance at temperature 0.6, top-$p$ 0.95, and top-$k$ 20. At exactly 512 Countdown groups and update 28, $4{:}2{:}1$ reaches 38.57\% and $4{:}1{:}2$ reaches 34.91\%, both with Countdown quota $4/7$, \rev{showing that companion-task allocation also changes matched-exposure returns at the 4B scale}.

\paragraph{Allocation diagnostics.}
Figure~\ref{fig:ratio-allocation-tradeoff} varies the Chemistry--Precalculus allocation and evaluates held-out performance after 25 pure-random updates. Dashed lines mark the common initial policy.

\par
\WFclear
\endgroup

\Needspace{6\baselineskip}
\begingroup
\subsection{Detailed Experimental Hyperparameters}

Table~\ref{tab:qel-hyperparameters} lists the TextArena QEL settings. QEL matches task exposure, whereas FEAST matches optimizer updates and rollout count.

\input{tables/qel_hyperparameters}

\input{tables/feast_hyperparameters}
\FloatBarrier
\endgroup
\endgroup

\subsection{\rev{Additional Analyses}}
\label{app:additional-analyses}
{\color{black}
This section collects the analyses added during the review. Unless stated otherwise, runs use GURU-18K with Qwen3-1.7B-Base, 100 updates, and 1,024 rollouts per update.

\subsubsection{Statistical significance and per-benchmark gains}
\label{app:add-significance}
Table~\ref{tab:add-significance} reports paired FEAST$-$Random differences. The question-level 95\% CIs come from a bootstrap over the 4,199 evaluation questions with 10,000 resamples, holding the evaluated training seeds fixed. The intervals lie above zero at both scales. Avg.\ weights the nine benchmarks equally, so large gains on individual benchmarks, such as FinQA at 1.7B and CodeIO at 8B, each add less than one point to Avg.\ (Table~\ref{tab:add-benchmark-gains}). At 8B, the Code gap comes mainly from HumanEval, while MBPP remains nearly unchanged. The lower Code rollout share in Figure~\ref{fig:guru-domain-allocation} is a possible explanation.

\begin{table}[!htbp]
\color{black}
\centering
\small
\caption{FEAST versus Random on GURU-18K. Scores are mean $\pm$ sample SD over training seeds.}
\label{tab:add-significance}
\begin{tabular}{lcccc}
\toprule
Model & Random & FEAST & FEAST $-$ Random & 95\% CI \\
\midrule
1.7B & 35.43 $\pm$ 0.18 & 36.18 $\pm$ 0.13 & $+0.75$ & [0.20, 1.31] \\
8B & 50.65 $\pm$ 1.01 & 51.69 $\pm$ 0.08 & $+1.04$ & [0.30, 1.79] \\
\bottomrule
\end{tabular}
\end{table}

\begin{table}[!htbp]
\color{black}
\centering
\small
\caption{Benchmarks with large FEAST gains. Gain is the FEAST$-$Random difference averaged over training seeds. Each benchmark contributes one ninth of its gain to Avg.}
\label{tab:add-benchmark-gains}
\begin{tabular}{llcc}
\toprule
Model & Benchmark & Gain & Contribution to Avg. \\
\midrule
1.7B & FinQA & $+7.08$ & $+0.79$ \\
8B & CodeIO & $+4.56$ & $+0.51$ \\
8B & MultiHierTT & $+3.61$ & $+0.40$ \\
\bottomrule
\end{tabular}
\end{table}

\subsubsection{Matched exposure on GURU reasoning data}
\label{app:add-qel-guru}
\textbf{Fixed Math quotas.} The varied factor is the Math share of each update, 12.5\% or 50\%. Both runs start from the same model and see the same 512 Math prompts in the same order, so the 12.5\% run reaches this exposure after 32 updates with more Code and Science training, and the 50\% run after 8 updates. The 12.5\% run scores higher on MATH-500 (Table~\ref{tab:add-math-quota}), so matched Math exposure does not determine the Math score. The direction is consistent with positive transfer from Code and Science, which the transfer term in Theorem~\ref{thm:qel-decomposition} allows.

\textbf{Matched Zebra exposure.} The varied factor is the original allocation history: Full FEAST versus FEAST w/o reserved budgets mechanism, which runs the later promotion over all selected prompts together (Appendix~\ref{app:experiment-details}). This history leaves Zebra with 3,360 rollouts under Full FEAST and 2,728 under FEAST w/o reserved budgets mechanism. We match the Zebra rollout count by giving FEAST w/o reserved budgets mechanism one extra update, then train both models for four updates on the same ordered Zebra, Code, and Science prompt stream with the same allocation. Each model keeps its own weights and optimizer state. At matched Zebra rollout counts, Full FEAST scores far higher (Table~\ref{tab:add-zebra}), so equal Zebra exposure does not give equal Zebra performance. Four further updates, each spending 512 of its 1,024 rollouts on Zebra, narrow the gap without closing it.

\begin{table}[!htbp]
\color{black}
\centering
\small
\caption{Same Math exposure under two allocation paths. Both runs train Qwen3-1.7B-Base on the same 512 Math prompts in the same order and differ only in the Math share of each update. MATH-500 uses avg@1. The 95\% CI comes from a paired question-level bootstrap.}
\label{tab:add-math-quota}
\begin{tabular}{lcccc}
\toprule
\begin{tabular}[c]{@{}l@{}}Math share\\per update\end{tabular} & \begin{tabular}[c]{@{}c@{}}Math\\prompts\end{tabular} & \begin{tabular}[c]{@{}c@{}}Code + Science\\prompts\end{tabular} & Updates & MATH-500 \\
\midrule
12.5\% & 512 & 3,584 & 32 & 61.20 \\
50\% & 512 & 512 & 8 & 55.40 \\
Difference (12.5\% $-$ 50\%), 95\% CI & N/A & N/A & N/A & $+5.80$ [2.00, 9.80] \\
\bottomrule
\end{tabular}
\end{table}

\begin{table}[!htbp]
\color{black}
\centering
\small
\caption{Zebra scores at matched cumulative rollout counts. The top-up update uses 79 Zebra and 49 background prompts. Each shared update uses 64 Zebra, 32 Code, and 32 Science prompts. Both stages generate eight responses per prompt. The 95\% CI comes from a paired question-level bootstrap.}
\label{tab:add-zebra}
\setlength{\tabcolsep}{5pt}
\begin{tabular}{lcccc}
\toprule
Stage & \begin{tabular}[c]{@{}c@{}}Zebra rollouts\\per model\end{tabular} & \begin{tabular}[c]{@{}c@{}}Full\\FEAST\end{tabular} & \begin{tabular}[c]{@{}c@{}}FEAST w/o reserved\\budgets mechanism\end{tabular} & Gap, 95\% CI \\
\midrule
After matching rollout counts & 3,360 & 31.87 & 8.78 & $+23.09$ [20.82, 25.49] \\
After two shared updates & 4,384 & 32.68 & 13.53 & $+19.15$ [16.92, 21.47] \\
After four shared updates & 5,408 & 33.54 & 17.41 & $+16.13$ [14.11, 18.18] \\
\bottomrule
\end{tabular}
\end{table}

\subsubsection{Comparison with the VIGOR allocation rule}
\label{app:add-vigor}
FEAST w/o inter-cluster fairness applies FEAST's [2,4,16] increments in a global pool, while VIGOR~\citep{jiang2026vigor} uses [2,4,8,16]. We therefore implement the published VIGOR allocation rule in the same GRPO framework, with the same normalization, zero-variance filtering, and 1,024 rollouts per update. FEAST achieves a higher Avg.\ (Table~\ref{tab:add-vigor}). The two methods share variance-guided allocation and differ in FEAST's cluster structure and reserved budgets, so this result supports the benefit of these additional mechanisms.

\begin{table}[!htbp]
\color{black}
\centering
\small
\caption{FEAST versus the VIGOR allocation rule at the same budget. FEAST allocates progressively inside four reserved budgets, and VIGOR over all 128 prompts.}
\label{tab:add-vigor}
\begin{tabular}{lccc}
\toprule
Method & Rollouts added per round & Prompts per round & Avg. \\
\midrule
Full FEAST & [2,4,16] & 32/16/8 in each of four clusters & 36.32 \\
VIGOR rule & [2,4,8,16] & 128/64/32/16 in one global pool & 35.58 \\
\bottomrule
\end{tabular}
\end{table}

\subsubsection{Initial pilot size}
\label{app:add-pilot}
A four-response initial pilot uses [4,4,8] increments in place of [2,4,16], keeping the total budget and the cluster allocation fixed, so only the within-cluster schedule changes. Its Avg.\ is close to Full FEAST (Table~\ref{tab:add-pilot}), suggesting that the noise of two-response variance estimates has little effect on performance. Giving every prompt eight responses in one round, the extreme case of a larger first round, lowers Avg., suggesting that short pilots free budget for concentrating rollouts on prompts with reward variation.

\begin{table}[!htbp]
\color{black}
\centering
\small
\caption{Rollout schedules at the same budget. Each update starts with 128 prompts. A schedule [a,b,c] gives a rollouts to every prompt, then b more to the half with the highest reward variance in each cluster, then c more to the top quarter.}
\label{tab:add-pilot}
\begin{tabular}{lccc}
\toprule
Method & Schedule & Rollouts per update & Avg. \\
\midrule
Full FEAST & [2,4,16] & 128$\times$2 + 64$\times$4 + 32$\times$16 = 1,024 & 36.32 \\
Four-response pilot & [4,4,8] & 128$\times$4 + 64$\times$4 + 32$\times$8 = 1,024 & 36.26 \\
FEAST w/o intra-cluster efficiency & [8] & 128$\times$8 = 1,024 & 34.66 \\
\bottomrule
\end{tabular}
\end{table}

\subsubsection{Atomic-unit service}
\label{app:add-service}
We compare Full FEAST with FEAST w/o exploration floor ($\epsilon=0$) on the same 64 atomic units. A unit is served in an update when it receives at least one prompt. Both runs serve every atomic unit. Compared with FEAST w/o exploration floor, FEAST substantially reduces the P90 and maximum waits (Table~\ref{tab:add-service}), indicating that the floor protects the units that would otherwise wait longest.

\begin{table}[!htbp]
\color{black}
\centering
\small
\caption{Atomic-unit coverage and waiting times over 100 updates on the same 64 units. Waiting statistics summarize each unit's longest unserved interval.}
\label{tab:add-service}
\begin{tabular}{lcc}
\toprule
Metric & Full FEAST & FEAST w/o exploration floor ($\epsilon=0$) \\
\midrule
Units served at least once & 64/64 & 64/64 \\
Longest wait: median / P90 / max (updates) & 10 / 20.7 / 59 & 8 / 52.5 / 73 \\
\bottomrule
\end{tabular}
\end{table}

}
\FloatBarrier

%% file: sec/appendix-qel.tex
\subsection{Theory for the Quota--Efficiency Law}
\label{app:qel-theory}
This section provides the formal setup and proofs for Section~\ref{sec:qel}.

\subsubsection{Setup and path decomposition}
\label{app:qel-setup}

Let $\mathcal F_t$ denote the training history before update $t$, and fix the initial parameters $\theta_0$. At step $t$, the scheduler allocates fractions $\mathbf q_t\in\Delta^{K-1}$ of a total batch budget $B_t>0$ across capabilities, with step size $\alpha_t\in[0,\alpha_{\max}]$ for finite $\alpha_{\max}$. Here $\theta_t,B_t,\mathbf q_t,\alpha_t$ are $\mathcal F_t$-measurable. Let $\widehat g_{j,t}$ be the square-integrable, $\mathcal F_{t+1}$-measurable update direction for capability $j$ with conditional mean $g_{j,t}=\mathbb E[\widehat g_{j,t}\mid\mathcal F_t]$. We study the first-order update model
\begin{equation}
\theta_{t+1}=\theta_t+\alpha_t\widehat g_t,
\qquad \widehat g_t=\sum_jq_{j,t}\widehat g_{j,t}.
\label{eq:qel-update}
\end{equation}

For schedule $\mathcal Q$, the cumulative exposure to capability $k$ and the stopping time to reach $N>0$ exposures are
\begin{equation}
N_k(T)=\sum_{t<T}B_tq_{k,t},\qquad
T=\tau_k^{\mathcal Q}(N)=\inf\{s:N_k(s)\geq N\}.
\label{eq:qel-stopping-time}
\end{equation}
Assume $N_k(T)=N$ almost surely and $T\leq T_{\max}<\infty$ for a deterministic bound $T_{\max}$. Further assume $V_k$ has an $L$-Lipschitz gradient and $\mathbb E[\|\widehat g_t\|^2\mid\mathcal F_t]\leq G^2$. Write $A_{k,t}=\langle\nabla V_k(\theta_t),g_{k,t}\rangle$ for the direct learning contribution and $C_{j\to k,t}=\langle\nabla V_k(\theta_t),g_{j,t}\rangle$ for transfer from capability $j$ to $k$.

\begin{theorem}[Quota-path decomposition]
\label{thm:qel-decomposition}
Under the setup above, the matched-exposure efficiency in \eqref{eq:qel-matched-exposure} satisfies
\begin{align}
\eta_k(N;\mathcal Q)
&=\frac1N\mathbb E\!\left[\sum_{t<T}\alpha_t\!\left(q_{k,t}A_{k,t}+\sum_{j\ne k}q_{j,t}C_{j\to k,t}\right)\right]+\mathcal R_N,
\label{eq:qel-decomposition}\\
|\mathcal R_N|&\leq\frac{LG^2}{2N}\mathbb E\!\left[\sum_{t<T}\alpha_t^2\right].
\label{eq:qel-remainder}
\end{align}
\end{theorem}

\begin{proof}
By $L$-smoothness of $V_k$,
\begin{equation}
V_k(\theta_{t+1})-V_k(\theta_t)
=\alpha_t\langle\nabla V_k(\theta_t),\widehat g_t\rangle+r_t,
\qquad |r_t|\leq\tfrac L2\alpha_t^2\|\widehat g_t\|^2.
\label{eq:qel-taylor-step}
\end{equation}
Multiplying by $\mathbf1\{t<T\}$, summing over $t<T_{\max}$, and taking expectations replaces each $\widehat g_{j,t}$ with $g_{j,t}$ by the tower property, since $\mathbf1\{t<T\}$, $\theta_t$, $\alpha_t$, and $\mathbf q_t$ are $\mathcal F_t$-measurable. Telescoping the left-hand side and dividing by $N$ yields \eqref{eq:qel-decomposition}, while the second-moment bound gives \eqref{eq:qel-remainder}.
\end{proof}

\subsubsection{Efficiency ordering under suppressive transfer}
\label{app:qel-staleness}

To isolate the effect of suppressive transfer, consider a readiness model for two deterministic schedules $\mathcal Q^+$ and $\mathcal Q^-$. Each schedule is partitioned into $m\geq1$ intervals of $\delta=N/m$ exposures to capability $k$. Let $h_{k,n}^{\mathcal Q}$ denote the off-capability exposure during interval $n$ (for a constant quota $q_k>0$, $h_{k,n}=\delta(1-q_k)/q_k$). Suppose $\mathcal Q^+$ incurs at most as much off-capability exposure as $\mathcal Q^-$ in every interval:
\begin{equation}
h_{k,n}^{\mathcal Q^+}\leq h_{k,n}^{\mathcal Q^-},\qquad n=1,\ldots,m.
\label{eq:qel-gap-dominance}
\end{equation}

We model the policy's readiness to learn $k$ as a scalar state $s_{k,n}$ evolving across intervals through shared maps
\begin{equation}
\widetilde s_{k,n}=\mathsf D_{k,h_{k,n}}(s_{k,n}),\qquad
s_{k,n+1}=\mathsf U_k(\widetilde s_{k,n}),
\label{eq:qel-readiness-dynamics}
\end{equation}
where the decay map $\mathsf D_{k,h}(s)$ is nondecreasing in $s$ and nonincreasing in $h$, and the update map $\mathsf U_k$, rollout quality $r_k$, and value map $v_k$ are nondecreasing. Let both schedules share the initial state $s_{k,1}$ with $V_k(\theta_0)=v_k(s_{k,1})$, and let $\mathbb E[V_k(\theta_T^{\mathcal Q})]=v_k(s_{k,m+1}^{\mathcal Q})$.

\begin{proposition}[QEL in a suppressive regime]
\label{prop:qel-staleness}
Under the readiness dynamics and intervalwise exposure order above, for all $n=1,\ldots,m$,
\begin{equation}
\widetilde s_{k,n}^{\mathcal Q^+}\geq\widetilde s_{k,n}^{\mathcal Q^-},\qquad
r_k(\widetilde s_{k,n}^{\mathcal Q^+})\geq r_k(\widetilde s_{k,n}^{\mathcal Q^-}),
\label{eq:qel-quality-order}
\end{equation}
and the matched-exposure efficiency satisfies $\eta_k(N;\mathcal Q^+)\geq\eta_k(N;\mathcal Q^-)$.
\end{proposition}

\begin{proof}
Given $s_{k,n}^+\geq s_{k,n}^-$, the monotonicity of $\mathsf D$ implies
\[
\mathsf D_{k,h_{k,n}^+}(s_{k,n}^+)
\geq\mathsf D_{k,h_{k,n}^+}(s_{k,n}^-)
\geq\mathsf D_{k,h_{k,n}^-}(s_{k,n}^-).
\]
Since $\mathsf U_k$ is nondecreasing, induction from $s_{k,1}^+=s_{k,1}^-$ establishes $s_{k,n}^+\geq s_{k,n}^-$ for all $n$. Applying the nondecreasing maps $r_k$ and $v_k$ yields \eqref{eq:qel-quality-order} and the final efficiency ordering.
\end{proof}

\paragraph{Strict inequality.}
The efficiency inequality is strict if $\mathsf D_{k,h_{k,n_0}^+}(s_{k,n_0}^-)>\mathsf D_{k,h_{k,n_0}^-}(s_{k,n_0}^-)$ for some interval $n_0$, and $\mathsf U_k$, $v_k$, and subsequent $\mathsf D_{k,h_{k,n}^+}(\cdot)$ are strictly increasing. For example, setting $\mathsf D_{k,h}(s)=e^{-\lambda h}s$, $\mathsf U_k(s)=s+\kappa s(1-s)$, and $r_k(s)=v_k(s)=s$ on $s\in(0,1]$ with $\lambda>0$ and $\kappa\in(0,1)$ gives strict efficiency ordering under \eqref{eq:qel-gap-dominance} whenever $h_{k,n_0}^+<h_{k,n_0}^-$ in one interval.

\subsubsection{Sufficient quota threshold}
\label{app:qel-critical}

\begin{corollary}[Effective-quota threshold for nonprogress]
\label{cor:qel-critical}
In the setting of Theorem~\ref{thm:qel-decomposition}, suppose $\alpha_t=\alpha>0$ and there exist constants $a,b>0$ such that almost surely for $t<T$,
\begin{equation}
A_{k,t}\leq a,\qquad
\sum_{j\ne k}q_{j,t}C_{j\to k,t}\leq-b(1-q_{k,t}).
\label{eq:qel-suppressive-transfer}
\end{equation}
Define the effective quota over the horizon $T$ as
\begin{equation}
q_{k,N}^{\mathrm{eff}}=
\frac{\mathbb E[\sum_{t<T}q_{k,t}]}{\mathbb E[T]},
\label{eq:qel-effective-quota}
\end{equation}
which simplifies to $N/(B\mathbb E[T])$ when $B_t\equiv B$, and to $q_k$ when $q_{k,t}\equiv q_k$. If $b>LG^2\alpha/2$, then
\begin{equation}
q_{k,N}^{\mathrm{eff}}\leq\frac{b-LG^2\alpha/2}{a+b}
\quad\Longrightarrow\quad\eta_k(N;\mathcal Q)\leq0.
\label{eq:qel-critical}
\end{equation}
As $\alpha\to0$ with $a,b,L,G$ fixed, this threshold approaches $b/(a+b)$.
\end{corollary}

\begin{proof}
Substituting \eqref{eq:qel-suppressive-transfer} and \eqref{eq:qel-remainder} into Theorem~\ref{thm:qel-decomposition} gives
\begin{equation}
\eta_k(N;\mathcal Q)
\leq\frac{\alpha\mathbb E[T]}{N}
\left[(a+b)q_{k,N}^{\mathrm{eff}}-b+\frac{L\alpha G^2}{2}\right],
\label{eq:qel-critical-bound}
\end{equation}
which is nonpositive under condition \eqref{eq:qel-critical}.
\end{proof}

\subsubsection{Learning signals under binary-reward GRPO}
\label{app:qel-grpo}

Consider a prompt $x$ under policy $\theta$, where $G_r\geq2$ independent rollouts receive Bernoulli rewards with success probability $p_x(\theta)$. The probability that a group contains both successes and failures is
\begin{equation}
\iota_{G_r}(p)=1-p^{G_r}-(1-p)^{G_r}.
\end{equation}
The function $\iota_{G_r}(p)$ increases on $[0,1/2]$ and decreases on $[1/2,1]$, with $\iota_{G_r}(p)=G_rp+O(p^2)$ near zero. For $G_r=2$, $\iota_2(p)=2p(1-p)$ recovers the two-pilot disagreement rate used by FEAST.

Under the zero-advantage convention~\citep{yu2025dapo}, groups with uniform rewards produce zero reward-centered task gradients. Let $a_{k,x}(\theta)$ denote the expected alignment of the task gradient with $\nabla V_k(\theta)$, and $\mu_{k,x}(\theta)$ denote this alignment conditioned on observing mixed outcomes. Then for $p_x(\theta)\in(0,1)$,
\begin{equation}
a_{k,x}(\theta)=\iota_{G_r}(p_x(\theta))\mu_{k,x}(\theta),
\label{eq:qel-grpo-contribution}
\end{equation}
and $a_{k,x}(\theta)=0$ at $p_x(\theta)\in\{0,1\}$. Consequently, in the low-success regime $0<p_x(\theta^-)\leq p_x(\theta^+)\leq1/2$, if $0\leq\mu_{k,x}(\theta^-)\leq\mu_{k,x}(\theta^+)$, monotonicity of both factors guarantees $a_{k,x}(\theta^+)\geq a_{k,x}(\theta^-)$. Averaging this pointwise ordering over a common prompt distribution preserves it.

\paragraph{Greedy starvation.}
\label{app:qel-starvation}
Suppose a greedy scheduler assigns $q_{k,t}=0$ whenever its priority score satisfies $z_{k,t}\leq0$, and receiving zero quota causes the score to stagnate or decay ($q_{k,t}=0\Rightarrow z_{k,t+1}\leq z_{k,t}$). Then any state with $z_{k,t_0}\leq0$ is absorbing, permanently starving capability $k$ ($q_{k,t}=0$ for all $t\geq t_0$).

Together, these results complement prior analyses of gradient imbalance~\citep{wu2025imbalanced}, worst-task return gaps~\citep{corrado2026drats}, and low-variance reward signals~\citep{wang2026ragen2} by characterizing how allocation history shapes per-sample efficiency.

%% file: sec/appendix-feast.tex
\subsection{FEAST Details and Fair-Service Analysis}
\label{app:feast-details}

Algorithm~\ref{alg:feast} summarizes FEAST's training loop. The following subsections specify its grouping, allocation, and feedback rules and prove the group-opportunity bound.

\begin{algorithm}[!htbp]
\caption{FEAST training}
\label{alg:feast}
\small
\textbf{Input:} Policy $\pi_{\theta_0}$, prompt pool $\mathcal X$, verifier, update horizon $T$.\par
\textbf{Budget:} $B$ rollouts per update, shared equally among $M$ selected clusters.\par\smallskip
\begin{tabularx}{\linewidth}{@{}r@{\quad}>{\raggedright\arraybackslash}X@{}}
1 & Initialize persistent atomic units, their histories, and capability clusters.\\
2 & \textbf{for} $t=0,\ldots,T-1$ \textbf{do}\\
3 & \quad Periodically regroup units using current-policy representations and learning trends.\\
4 & \quad Compute positive cluster weights from atomic histories (Eqs.~\ref{eq:feast-atomic-score}--\ref{eq:feast-group-score}).\\
5 & \quad Sample $M$ clusters without replacement; reserve $B_c=B/M$ rollouts for each.\\
6 & \quad $\mathcal D_t\gets\varnothing$; hold $\pi_{\theta_t}$ fixed during generation.\\
7 & \quad \textbf{for each} selected cluster $c$ \textbf{do}\\
8 & \qquad Sample prompts using atomic priorities and remaining prompt capacity.\\
9 & \qquad Generate and score an equal number of pilot rollouts per prompt.\\
10 & \qquad Spend the remaining rollouts in rounds, retaining prompts by cumulative reward variance.\\
11 & \qquad Add all $B_c$ rollouts to $\mathcal D_t$, grouped by prompt.\\
12 & \quad \textbf{end for}\\
13 & \quad $\theta_{t+1}\gets\operatorname{PolicyUpdate}(\theta_t,\text{signal-bearing prompt groups in }\mathcal D_t)$.\\
14 & \quad Refresh pilot-based attainment and productivity, learning trends, and rollout shortfalls.\\
15 & \textbf{end for}\\
16 & \textbf{return} $\pi_{\theta_T}$\\
\end{tabularx}
\end{algorithm}

\subsubsection{Persistent units and policy-adaptive groups}

We embed the unlabeled training prompts with the base policy and partition them into $J=64$ atomic units. Each prompt is rendered with the model's chat template, including the assistant-generation prefix, and truncated to the training input cap (2,048 tokens for GURU-18K). We use the final transformer hidden states after the model's final normalization and before the language-model head. The prompt vector is the attention-mask-weighted mean over all non-padding input tokens, including template tokens; no response tokens or $\ell_2$ normalization are used. Pooling accumulates in float32. The initial partition fits Euclidean K-means to these raw prompt vectors, using K-means++ initialization, Lloyd iterations, ten restarts, at most 300 iterations, tolerance $10^{-4}$. Each prompt keeps its nearest-centroid atomic assignment throughout training; the initial partition is not recomputed after input-length filtering.

Every ten optimizer updates, the current policy re-embeds the retained training pool using the same hidden layer and pooling rule. We take an equal-weight mean of prompt vectors within each atomic unit. The retained pool consists of prompts eligible for training after length filtering and need not contain every prompt used to construct the initial partition. Let $e_{i,t}$ be the resulting centroid. We center the 64 centroids, project them onto the leading eight right-singular vectors, and divide each projected coordinate by its empirical standard deviation, obtaining $z_{i,t}\in\mathbb R^8$. The regrouping feature is $[z_{i,t},\sqrt{\lambda_y}\,\bar\ell_{i,t}]$, where $\bar\ell_{i,t}$ is a slowly smoothed signed progress signal and $\lambda_y=1$.

For each component count from 8 through 16, we fit a diagonal-covariance Gaussian mixture and compute its BIC. We then repair assignments, using log posterior responsibilities to avoid probability underflow, to require at least two atomic units per group. Among feasible repaired assignments, we select the component count with the smallest BIC of its original fitted mixture. Regrouping changes only the map $i\mapsto c_t(i)$, while all attainment, productivity, and exposure statistics remain indexed by atomic unit.

The states $m,u,\ell,\bar\ell$ are initialized to zero and remain unchanged for atoms that are not sampled. Pilot feedback uses the common two-rollout first round. For each served prompt $x$, let $R_{x,j}\in[0,1]$ be the raw verifier reward on pilot $j$, preserving partial credit. The attainment observation is the prompt-equal mean $\widehat m=|P|^{-1}\sum_{x\in P}(R_{x,1}+R_{x,2})/2$, and the productive-signal observation is $\widehat u=|P|^{-1}\sum_{x\in P}(R_{x,1}-R_{x,2})^2$. With binary rewards the latter equals the disagreement indicator, while for partial rewards it retains the magnitude of the difference. The first observation sets $m$ and $u$ to $\widehat m$ and $\widehat u$; subsequent observations use $m_{i,t+1}=0.9m_{i,t}+0.1\widehat m$ and $u_{i,t+1}=0.9u_{i,t}+0.1\widehat u$. Promotion and optimizer-signal detection also use the original verifier scores.

Let $n^{\mathrm{sig}}_{i,t}$ be the number of generated trajectories in the served batch belonging to prompt groups with nonzero verifier-score variance. The instantaneous signed trend and the slower regrouping trend are
\begin{equation}
\ell_{i,t+1}=\operatorname{clip}\!\left(\frac{m_{i,t+1}-m_{i,t}}{n^{\mathrm{sig}}_{i,t}/8},-1,1\right),\qquad \bar\ell_{i,t+1}=0.7\bar\ell_{i,t}+0.3\ell_{i,t+1},
\label{eq:feast-progress-update}
\end{equation}
with $\ell_{i,t+1}=0$ on the first observation or when $n^{\mathrm{sig}}_{i,t}=0$. FEAST tracks each atom's accumulated rollout shortfall relative to an equal per-update share. With $J=64$ atoms, equal-share increment $b=B/J=16$, and all-round generated trajectory count $g_{i,t}$, it updates
\begin{equation}
d_{i,0}=0,\qquad d_{i,t+1}=\max\{0,d_{i,t}+b-g_{i,t}\},\qquad
a_{i,t}=\min\{1,d_{i,t}/(10b)\}.
\end{equation}
Shortfalls are updated for all atoms, counting zero-variance trajectories as exposure.

\subsubsection{Cluster priorities and prompt quotas}

The atomic priority is
\begin{equation}
r_{i,t}=\epsilon+u_{i,t}\,\phi(m_{i,t})\,\psi(m_{i,t})\bigl(1+\gamma|\ell_{i,t}|\bigr)+\lambda_a a_{i,t},
\label{eq:feast-atomic-score}
\end{equation}
where
\begin{equation}
\phi(m)=\exp\!\left[-\frac12\left(\frac{m-1/2}{\sigma}\right)^2\right],\qquad \psi(m)=1+\frac{\eta}{1+\exp((m-1/2)/\tau)}.
\label{eq:feast-score-factors}
\end{equation}
The factor $\Phi$ in \eqref{eq:feast-priority} is $\Phi(m)=\phi(m)\psi(m)$. The resolved values are $\epsilon=0.05$, $\sigma=0.22$, $\eta=0.20$, $\tau=0.05$, $\gamma=0.50$, and $\lambda_a=0.05$. Because $u,m,a\in[0,1]$ and $|\ell|\leq1$, the current instantiation obeys $0.05\leq r_{i,t}\leq1.90$.

Let $n^{\mathrm{unseen}}_{i,t}$ count prompts in atom $i$ that have never been sampled during training, let $P=32$ be the prompt budget of a selected group, and let $U_{c,t}=\sum_{i:c_t(i)=c}n^{\mathrm{unseen}}_{i,t}$. For $U_{c,t}>0$, define the capacity-weighted atomic mean $\bar r_{c,t}=U_{c,t}^{-1}\sum_{i:c_t(i)=c}n^{\mathrm{unseen}}_{i,t}r_{i,t}$. If $U_{c,t}=0$, use the unweighted atomic mean. As lifetime-unseen capacity is exhausted, the group score shrinks toward the exploration floor according to
\begin{equation}
R_{c,t}=\epsilon+\min\!\left\{1,\frac{U_{c,t}}{P}\right\}(\bar r_{c,t}-\epsilon),\qquad \widetilde w_{c,t}=R_{c,t}^{1/\mathcal T}.
\label{eq:feast-group-score}
\end{equation}
Here $R_{c,t}=w_{c,t}$ in the main-text notation. FEAST samples four groups without replacement using the next-draw probabilities in \eqref{eq:feast-inter}. Within a selected group, each of its 32 prompt slots is assigned sequentially to an atomic unit with probability proportional to its priority times remaining lifetime-unseen capacity. After this capacity is exhausted, the fallback uses each atomic unit's total eligible prompt capacity minus the slots already assigned to it in the current batch. Prompt identity selection is handled separately through stable per-atom without-replacement cycles, with repeats admitted after cycle exhaustion.

\subsubsection{Fair-service bound}
\label{app:feast-fairness-proof}

\begin{proposition}[Non-absorbing group opportunity]
\label{prop:feast-service}
Consider an indefinitely continued FEAST process. Suppose that at every update the current group containing a persistent atomic unit $i$ is eligible, there are at most $C_{\max}$ eligible groups, and the temperature-scaled sampling weights satisfy $0<\underline w\leq\widetilde w_{c,t}\leq\overline w<\infty$ almost surely. FEAST samples at least one group without replacement. Let $W_i^{\mathrm{grp}}$ count future updates through and including the first selection of the then-current group containing $i$, with $W_i^{\mathrm{grp}}=\infty$ if no such selection occurs. Conditional on any training history, with $p_{\min}=\underline w/(C_{\max}\overline w)$ and integer $h\geq0$,
\begin{equation}
\Pr(W_i^{\mathrm{grp}}>h\mid\mathcal F_t)\leq(1-p_{\min})^h,\qquad \mathbb E[W_i^{\mathrm{grp}}\mid\mathcal F_t]\leq p_{\min}^{-1}.
\label{eq:feast-fairness}
\end{equation}
\end{proposition}

\begin{proof}
At every update, even the first draw selects the current group containing $i$ with probability at least $\underline w/(C_{\max}\overline w)=p_{\min}$. Successive conditioning therefore bounds the probability of $h$ consecutive misses by $(1-p_{\min})^h$, without requiring independence across updates. Summing these conditional tail bounds gives $\mathbb E[W_i^{\mathrm{grp}}\mid\mathcal F_t]\leq\sum_{h\geq0}(1-p_{\min})^h=p_{\min}^{-1}$.
\end{proof}

For the score above, one may take $\underline w=0.05^{1/\mathcal T}$ and $\overline w=1.90^{1/\mathcal T}$. With $\mathcal T=1$ and $C_{\max}=16$, the first-draw bound gives $p_{\min}=1/608$.

\subsubsection{Budget-preserving progressive allocation}
\label{app:feast-progressive-budget}

For every selected group, the current schedule starts from 32 prompts with two rollouts each, retains the top half by cumulative verifier-score variance, adds four rollouts to each survivor, retains the top half again, and adds sixteen rollouts to each final survivor. Thus one group uses $32\times2+16\times4+8\times16=256$ trajectories. Four selected groups use exactly 1,024 trajectories, equal to assigning eight rollouts uniformly to 128 prompts. The final prompt-group sizes are 64 groups of size 2, 32 of size 6, and 32 of size 22. Stable tie-breaking is used when variance scores agree. The policy is held fixed across the within-update rounds, and all generated trajectories are associated with their original prompt before the reward-centered policy update. Prompt groups with zero verifier-score variance contribute no reward-centered optimizer signal, although an implementation may retain the minimum set of such groups solely for data-parallel alignment. All generated trajectories still count against the rollout budget.

\input{sec/appendix-feast-hyperparameters}

\subsubsection{\rev{One FEAST update}}
\label{app:feast-walkthrough}
{\color{black}
Table~\ref{tab:feast-walkthrough} follows one update in the main configuration (64 atomic units, four selected clusters, and 1,024 rollouts) and maps each step to the method.

\begin{table}[!htbp]
\color{black}
\centering
\small
\caption{One FEAST update in the main configuration, with the matching parts of the method.}
\label{tab:feast-walkthrough}
\begin{tabularx}{\linewidth}{@{}lXl@{}}
\toprule
Step & What FEAST does & Reference \\
\midrule
Retain histories & Keep 64 persistent atomic units and their states & Section~\ref{sec:method/clustering} \\
Regroup & Every 10 updates, regroup units by current-policy representations and smoothed learning trends & Eq.~\ref{eq:feast-grouping} \\
Compute priorities & Combine the exploration floor, rollout shortfall, and learning signals & Eq.~\ref{eq:feast-priority} \\
Select and reserve & Select four clusters without replacement and reserve 256 rollouts for each & Eqs.~\ref{eq:feast-budget}, \ref{eq:feast-inter} \\
Allocate progressively & In each cluster, give 2 rollouts to 32 prompts, 4 more to the top 16, and 16 more to the top 8 & Eq.~\ref{eq:feast-intra} \\
Update the policy & Train on 128 prompt groups: 64 with 2 responses, 32 with 6, and 32 with 22 & Eq.~\ref{eq:feast-loss}, Alg.~\ref{alg:feast} \\
Update feedback & Update unit histories. For example, pilot rewards 0 and 1 give attainment 0.5 and squared disagreement 1 & Appendix~\ref{app:feast-details} \\
\bottomrule
\end{tabularx}
\end{table}

\textbf{Policy loss.} GRPO normalizes rewards within each prompt's full response group, whether it has 2, 6, or 22 responses. With group mean $\bar R_x$ and sample standard deviation $s_x$, the advantage is $A_{xi}=(R_{xi}-\bar R_x)/(s_x+\epsilon_{\mathrm{adv}})$ for a small stabilizer $\epsilon_{\mathrm{adv}}$. For valid responses $\mathcal B$ in an optimization minibatch, the policy loss averages tokens within each response and responses within the minibatch,
\begin{equation}
L_{\mathcal B}(\theta)=-\frac{1}{|\mathcal B|}\sum_{(x,i)\in\mathcal B}\frac{1}{T_{xi}}\sum_{t=1}^{T_{xi}}\ell^{\mathrm{clip}}_{xit}(\theta),
\label{eq:feast-loss}
\end{equation}
where $\ell^{\mathrm{clip}}$ is the clipped GRPO surrogate and $T_{xi}$ counts valid response tokens. Each retained response contributes one term, so promoted prompts receive more weight in the update.
}

\FloatBarrier

%% file: sec/appendix-feast-hyperparameters.tex
\subsubsection{Hyperparameter design rationale}
\label{app:feast-hyperparameter-rationale}

Table~\ref{tab:feast-hyperparameters} reports the default settings, which were fixed before the ablation studies. We summarize their theoretical and structural rationale below.

\paragraph{Atomic resolution.}
The atomic-unit count $J$ controls the granularity of persistent statistics. Larger $J$ permits finer tracking but requires more histories and centroids to maintain and regroup. With $J=64$ and 8--16 dynamic groups, each group contains 4--8 atomic units on average, separating the resolution of persistent tracking from that of group-level allocation.

\paragraph{Exposure compensation.}
The positive priority floor $\epsilon$ supplies the lower bound used in Appendix~\ref{app:feast-fairness-proof}. The term $\lambda_a a_{i,t}$ adds a bounded correction for accumulated rollout shortfall. Since $a_{i,t}\in[0,1]$, setting $\lambda_a=\epsilon=0.05$ bounds this correction by the priority floor while allowing priority to increase with shortfall.

\paragraph{Pilot disagreement and the attainment frontier.}
For two conditionally independent rollouts of a fixed prompt under the same policy,
\begin{equation}
\mathbb E[(R_1-R_2)^2\mid x]=2\operatorname{Var}(R\mid x).
\label{eq:feast-pilot-variance-rationale}
\end{equation}
With Bernoulli reward of success probability $p$, this becomes $2p(1-p)$, maximized at $p=1/2$. This supports using pilot disagreement as a variance signal and motivates the moderate-attainment frontier, also used as a target in BOTS~\citep{shen2026bots}.

\paragraph{Absolute progress.}
Using $|\ell|$ allows both improvement and regression to increase priority. ALP-GMM~\citep{portelas2020teacher} similarly uses absolute learning progress to respond to improving and deteriorating competence.

\paragraph{Progressive rollout allocation.}
VIGOR~\citep{jiang2026vigor} provides a precedent for two initial rollouts and repeated top-half, variance-guided promotion. Appendix~\ref{app:feast-progressive-budget} gives the budget accounting for FEAST's three-round schedule.

%% file: tables/guru18k_evaluation.tex
\begin{wraptable}{R}{0.52\textwidth}
\vspace{-\dimexpr\intextsep+\abovecaptionskip\relax}
\centering
\normalsize
\setlength{\belowcaptionskip}{\baselineskip}
\setlength{\tabcolsep}{3pt}
\caption{GURU-18K evaluation configuration.}
\label{tab:guru18k-evaluation}
\begin{tabular*}{\linewidth}{@{\extracolsep{\fill}}lrrr@{}}
\toprule
Benchmark & Questions & $k$ & Cap \\
\midrule
MATH-500 & 500 & 1 & 4,096 \\
HumanEval & 164 & 4 & 4,096 \\
MBPP & 500 & 1 & 4,096 \\
GPQA-Diamond & 198 & 4 & 4,096 \\
SuperGPQA & 1,000 & 1 & 4,096 \\
Zebra & 200 & 4 & 8,192 \\
CodeIO & 200 & 4 & 8,192 \\
FinQA & 1,101 & 1 & 8,192 \\
MultiHierTT & 336 & 4 & 8,192 \\
\midrule
AMC & 83 & 4 & 8,192 \\
\bottomrule
\end{tabular*}
\par\smallskip
\raggedright $k$ denotes responses per question, and cap is the maximum number of response tokens.
\end{wraptable}

%% file: tables/feast_benchmarks.tex
\begingroup
\normalsize
\setlength{\tabcolsep}{1.6pt}
\begin{longtable}{@{}>{\raggedright\arraybackslash}p{1.15in}*{10}{r}@{}}
\caption{Individual benchmark scores (\%) for Qwen3-1.7B-Base, grouped by training seed. Trained checkpoints are at update 100, except ODF at update 50. Base is the untrained checkpoint at update 0. AMC uses an 8,192-token response cap and is excluded from the main-table Avg.}\label{tab:guru18k-seeds}\\
\toprule
Method & MATH & AMC & H.Eval & MBPP & GPQA & S.GPQA & Zebra & CodeIO & FinQA & M.Hier \\
\midrule
\endfirsthead
\multicolumn{11}{c}{Table~\thetable{} (continued). Qwen3-1.7B-Base.} \\
\toprule
Method & MATH & AMC & H.Eval & MBPP & GPQA & S.GPQA & Zebra & CodeIO & FinQA & M.Hier \\
\midrule
\endhead
\midrule
\multicolumn{11}{r}{\textit{Continued on the next page.}} \\
\endfoot
\bottomrule
\endlastfoot
Base & 51.40 & 27.41 & 46.19 & 52.40 & 15.03 & 12.30 & 1.46 & 2.88 & 8.90 & 13.47 \\
\midrule
\multicolumn{11}{@{}l}{\textit{Training seed 42}} \\*
Random & 62.00 & 27.41 & 69.51 & 55.80 & 30.43 & 21.80 & 34.90 & 4.00 & 16.44 & 23.81 \\
TAC & 64.60 & 28.31 & 67.53 & 53.40 & 29.67 & 21.70 & 23.97 & 5.50 & 23.07 & 27.60 \\
ADS & 60.60 & 30.12 & 68.90 & 54.00 & 33.08 & 24.20 & 0.38 & 3.75 & 3.54 & 11.09 \\
DUMP & 57.80 & 28.61 & 68.29 & 54.00 & 32.58 & 21.00 & 34.63 & 5.00 & 19.16 & 27.23 \\
BOTS & 61.20 & 33.73 & 68.14 & 54.40 & 30.56 & 22.70 & 3.60 & 5.38 & 23.98 & 27.23 \\
\textbf{FEAST} & 61.80 & 30.42 & 68.90 & 53.80 & 31.69 & 21.40 & 29.50 & 5.00 & 26.98 & 27.83 \\
DAPO & 58.00 & 28.61 & 69.51 & 55.60 & 30.18 & 23.70 & 33.60 & 4.38 & 23.16 & 27.60 \\
DAPO+FEAST & 63.00 & 33.43 & 70.12 & 57.60 & 32.07 & 23.60 & 35.32 & 4.25 & 26.16 & 32.07 \\
ODF & 58.40 & 29.52 & 68.29 & 56.20 & 32.58 & 21.20 & 1.13 & 3.75 & 17.89 & 27.83 \\
FEAST w/o inter-cluster fairness & 57.80 & 29.82 & 66.92 & 53.80 & 29.29 & 24.50 & 33.74 & 5.25 & 22.62 & 27.01 \\
FEAST w/o intra-cluster efficiency & 61.00 & 26.20 & 67.23 & 57.40 & 30.43 & 24.40 & 19.98 & 5.12 & 20.98 & 25.37 \\
\midrule
\multicolumn{11}{@{}l}{\textit{Training seed 43}} \\*
Random & 60.60 & 29.22 & 67.99 & 55.00 & 32.95 & 23.00 & 34.26 & 4.12 & 18.17 & 24.48 \\
TAC & 63.80 & 29.82 & 67.53 & 55.00 & 30.05 & 22.40 & 28.10 & 4.00 & 25.07 & 28.35 \\
ADS & 57.00 & 24.10 & 65.85 & 52.40 & 31.44 & 24.40 & 1.34 & 4.25 & 5.54 & 11.76 \\
DUMP & 63.60 & 29.52 & 68.75 & 56.00 & 30.05 & 22.80 & 8.41 & 4.88 & 21.62 & 27.75 \\
BOTS & 62.40 & 27.11 & 70.12 & 54.20 & 33.59 & 21.80 & 3.61 & 4.88 & 21.16 & 26.86 \\
\textbf{FEAST} & 57.80 & 27.11 & 68.29 & 55.20 & 33.59 & 23.50 & 32.15 & 5.00 & 23.52 & 26.26 \\
DAPO & 62.60 & 31.33 & 68.45 & 54.40 & 36.99 & 22.60 & 33.46 & 3.88 & 23.07 & 29.24 \\
DAPO+FEAST & 60.00 & 31.33 & 70.27 & 54.60 & 34.72 & 22.50 & 34.40 & 3.62 & 26.70 & 31.99 \\
\midrule
\multicolumn{11}{@{}l}{\textit{Training seed 44}} \\*
Random & 60.40 & 28.92 & 67.99 & 54.80 & 31.31 & 21.60 & 33.58 & 4.62 & 17.08 & 25.97 \\
TAC & 64.00 & 30.42 & 66.01 & 55.00 & 27.65 & 21.30 & 14.09 & 5.12 & 23.34 & 27.23 \\
ADS & 57.20 & 27.11 & 65.70 & 53.20 & 33.08 & 23.10 & 2.64 & 4.75 & 4.45 & 11.38 \\
DUMP & 60.40 & 30.42 & 68.60 & 54.60 & 30.18 & 23.70 & 32.90 & 6.25 & 16.62 & 23.96 \\
BOTS & 64.80 & 28.01 & 68.60 & 55.80 & 33.46 & 23.00 & 6.55 & 4.50 & 20.71 & 25.00 \\
\textbf{FEAST} & 61.60 & 31.63 & 68.45 & 56.20 & 28.91 & 22.90 & 33.36 & 4.38 & 22.43 & 26.41 \\
DAPO & 60.00 & 30.12 & 69.66 & 56.20 & 34.72 & 22.50 & 33.64 & 4.50 & 23.61 & 28.42 \\
DAPO+FEAST & 61.00 & 31.33 & 71.49 & 55.60 & 31.94 & 24.00 & 33.79 & 4.12 & 25.16 & 29.76 \\
\end{longtable}
\endgroup

%% file: tables/feast_benchmarks_8b.tex
\begingroup
\normalsize
\setlength{\tabcolsep}{2.2pt}
\begin{longtable}{@{}>{\raggedright\arraybackslash}p{0.98in}*{10}{r}@{}}
\caption{Individual benchmark scores (\%) for Qwen3-8B-Base, grouped by training seed. GRPO schedulers use update 100; DAPO and DAPO+FEAST use update 75. Base is the untrained checkpoint at update 0. AMC uses an 8,192-token response cap and is excluded from the main-table Avg.}\label{tab:guru18k-seeds-8b}\\
\toprule
Method & MATH & AMC & H.Eval & MBPP & GPQA & S.GPQA & Zebra & CodeIO & FinQA & M.Hier \\
\midrule
\endfirsthead
\multicolumn{11}{c}{Table~\thetable{} (continued). Qwen3-8B-Base.} \\
\toprule
Method & MATH & AMC & H.Eval & MBPP & GPQA & S.GPQA & Zebra & CodeIO & FinQA & M.Hier \\
\midrule
\endhead
\midrule
\multicolumn{11}{r}{\textit{Continued on the next page.}} \\
\endfoot
\bottomrule
\endlastfoot
Base & 73.00 & 42.47 & 84.76 & 69.40 & 38.76 & 29.30 & 2.77 & 8.50 & 28.70 & 35.34 \\
\midrule
\multicolumn{11}{@{}l}{\textit{Training seed 42}} \\*
Random & 78.80 & 56.93 & 88.87 & 73.20 & 42.93 & 34.20 & 34.65 & 5.50 & 41.69 & 49.55 \\
TAC & 78.20 & 51.81 & 87.96 & 73.60 & 44.57 & 34.50 & 38.50 & 10.38 & 39.87 & 50.67 \\
ADS & 78.80 & 54.22 & 85.82 & 74.00 & 48.48 & 34.90 & 16.65 & 6.25 & 41.24 & 47.99 \\
DUMP & 78.80 & 52.11 & 87.04 & 74.20 & 43.94 & 34.00 & 38.25 & 5.25 & 41.14 & 49.93 \\
BOTS & 81.20 & 54.52 & 86.74 & 73.40 & 46.84 & 35.60 & 28.11 & 6.00 & 41.51 & 52.75 \\
\textbf{FEAST} & 80.60 & 54.82 & 86.13 & 73.40 & 47.73 & 35.40 & 38.53 & 8.62 & 41.51 & 53.79 \\
DAPO & 76.40 & 53.61 & 86.89 & 75.00 & 42.55 & 35.40 & 38.40 & 7.50 & 38.87 & 50.22 \\
DAPO+FEAST & 79.20 & 55.12 & 89.02 & 72.60 & 46.09 & 36.80 & 38.77 & 5.88 & 40.78 & 52.31 \\
\midrule
\multicolumn{11}{@{}l}{\textit{Training seed 43}} \\*
Random & 82.00 & 50.30 & 87.96 & 73.20 & 47.73 & 36.60 & 35.83 & 7.25 & 41.14 & 50.52 \\
TAC & 79.60 & 54.22 & 87.96 & 74.20 & 43.56 & 37.50 & 35.97 & 5.62 & 41.87 & 51.26 \\
ADS & 77.80 & 52.41 & 88.57 & 73.00 & 47.10 & 35.40 & 14.78 & 5.62 & 34.15 & 42.56 \\
DUMP & 79.80 & 52.11 & 85.06 & 74.20 & 47.35 & 35.50 & 37.17 & 6.50 & 42.60 & 51.19 \\
BOTS & 78.80 & 47.59 & 88.26 & 73.80 & 46.09 & 34.30 & 28.95 & 9.12 & 42.78 & 50.74 \\
\textbf{FEAST} & 80.20 & 51.51 & 86.43 & 72.80 & 43.56 & 36.00 & 37.60 & 13.25 & 41.42 & 53.50 \\
DAPO & 79.60 & 55.12 & 88.11 & 73.20 & 43.43 & 36.50 & 38.69 & 6.12 & 41.14 & 52.60 \\
DAPO+FEAST & 77.80 & 52.71 & 88.41 & 75.00 & 45.83 & 34.30 & 38.26 & 6.62 & 40.87 & 51.41 \\
\end{longtable}
\endgroup

%% file: tables/feast_odf.tex
\begin{wraptable}{R}{0.38\textwidth}
\centering
\normalsize
\setlength{\belowcaptionskip}{\baselineskip}
\setlength{\tabcolsep}{4pt}
\setlength{\abovecaptionskip}{0pt}
\renewcommand{\arraystretch}{1.08}
\caption{ODF performance (\%) and generation cost. Results use 1.7B with seed 42 at update 50.}
\label{tab:guru18k-odf}
\begin{tabular*}{\linewidth}{@{\extracolsep{\fill}}lr@{}}
\toprule
Metric & 1.7B \\
\midrule
Math & \mbox{58.40} \\
Code & \mbox{62.25} \\
Science & \mbox{26.89} \\
Logic & \mbox{1.13} \\
Simulation & \mbox{3.75} \\
Tabular & \mbox{22.86} \\
Avg. & \mbox{31.92} \\
\midrule
Rollouts (k) & 841.7 \\
Tokens (M) & 737.5 \\
\bottomrule
\end{tabular*}
\end{wraptable}

%% file: tables/feast_computation.tex
\begin{table}[!htbp]
\centering
\normalsize
\setlength{\belowcaptionskip}{\baselineskip}
\caption{Online computation on GURU-18K (training seed 42). GRPO schedulers use 100 updates at both scales; DAPO and DAPO+FEAST use 100 at 1.7B and 75 at 8B. BOTS additionally uses a shared offline preparation artifact: 336.0k rollouts and 1,149.8M tokens.}
\label{tab:guru18k-computation}
\begin{tabular*}{\linewidth}{@{\extracolsep{\fill}}lrrrr@{}}
\toprule
& \multicolumn{2}{c}{Qwen3-1.7B-Base} & \multicolumn{2}{c}{Qwen3-8B-Base} \\
\cmidrule(lr){2-3}\cmidrule(lr){4-5}
Method & Rollouts (k) & Tokens (M) & Rollouts (k) & Tokens (M) \\
\midrule
Base & 0 & 0 & 0 & 0 \\
Random & 102.4 & 91.3 & 102.4 & 110.9 \\
ADS & 102.4 & 74.5 & 102.4 & 101.1 \\
BOTS & 102.4 & 92.3 & 102.4 & 108.8 \\
DUMP & 102.4 & 94.3 & 102.4 & 115.2 \\
TAC & 102.4 & 107.6 & 102.4 & 107.3 \\
\textbf{FEAST} & 102.4 & 101.4 & 102.4 & 99.8 \\
\midrule
DAPO & 426.0 & 357.8 & 205.8 & 180.1 \\
DAPO+FEAST & 399.4 & 373.5 & 206.8 & 201.0 \\
\bottomrule
\end{tabular*}
\end{table}

%% file: tables/feast_atomic_units.tex
\begin{wraptable}[7]{r}{0.40\textwidth}
\vspace{-\dimexpr\intextsep+\abovecaptionskip\relax}
\centering
\caption{Atomic-unit count, with $\Delta$ relative to the default $J=64$.}
\label{tab:guru18k-atomic-units}
\small
\setlength{\tabcolsep}{3pt}
\begin{tabular*}{\linewidth}{@{\extracolsep{\fill}}lrr@{}}
\toprule
Atomic units $J$ & Avg. & $\Delta$ \\
\midrule
32 & 33.84 & $-2.48$ \\
64 (default) & 36.32 & $N/A$ \\
128 & 36.39 & $+0.07$ \\
\bottomrule
\end{tabular*}
\end{wraptable}

%% file: tables/smc_domains.tex
\begin{table}[!htbp]
\centering
\normalsize
\setlength{\belowcaptionskip}{\baselineskip}
\setlength{\tabcolsep}{3pt}
\caption{SMC results by domain (avg@8). Avg.\ averages all 15 categories. Endpoints are update 100 at 1.7B and update 75 at 8B. Green and red mark the highest and lowest displayed scores within each scale, including ties.}
\label{tab:smc-domains}
\begin{tabular*}{\linewidth}{@{\extracolsep{\fill}}l*{8}{r}@{}}
\toprule
& \multicolumn{4}{c}{Qwen3-1.7B-Base} & \multicolumn{4}{c}{Qwen3-8B-Base} \\
\cmidrule(lr){2-5}\cmidrule(lr){6-9}
Method & Math & Code & Science & Avg. & Math & Code & Science & Avg. \\
\midrule
Random & 57.44 & \textcolor{resultworst}{\textbf{10.00}} & 72.74 & 61.42 & 72.50 & 32.50 & 90.89 & 78.42 \\
ADS & \textcolor{resultbest}{\textbf{57.50}} & 10.42 & \textcolor{resultworst}{\textbf{67.14}} & \textcolor{resultworst}{\textbf{58.86}} & \textcolor{resultworst}{\textbf{71.96}} & \textcolor{resultworst}{\textbf{27.50}} & \textcolor{resultworst}{\textbf{89.52}} & \textcolor{resultworst}{\textbf{77.19}} \\
DUMP & \textcolor{resultworst}{\textbf{55.71}} & \textcolor{resultbest}{\textbf{17.08}} & 77.86 & 63.47 & 72.20 & 33.75 & 91.73 & 78.75 \\
\textbf{FEAST} & 56.67 & \textcolor{resultbest}{\textbf{17.08}} & \textcolor{resultbest}{\textbf{79.64}} & \textcolor{resultbest}{\textbf{64.75}} & \textcolor{resultbest}{\textbf{74.29}} & \textcolor{resultbest}{\textbf{43.75}} & \textcolor{resultbest}{\textbf{92.38}} & \textcolor{resultbest}{\textbf{80.69}} \\
\bottomrule
\end{tabular*}
\end{table}

%% file: figures/sec_greedy_starvation.tex
\begin{figure}[!b]
\centering
\includegraphics[width=0.92\linewidth]{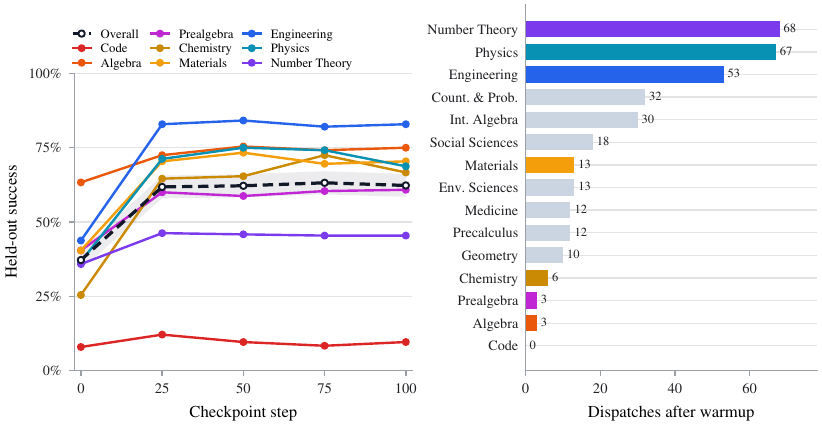}
\caption{SEC-Greedy, a deterministic Top-4 variant of Self-Evolving Curriculum (SEC), on SMC (Qwen3-1.7B). The left panel shows held-out success for the overall panel and selected clusters. The right panel shows class allocation after the 15-step balanced warmup. Code is never selected after warmup, while Algebra and Prealgebra receive only three dispatches each. Colored bars match the left-panel clusters. }
\label{fig:sec-greedy-starvation}
\end{figure}

%% file: sec/appendix-domain-trajectories.tex
\Needspace{7\baselineskip}
\begingroup
\setlength{\intextsep}{6pt}
\subsection{Domain-Level Allocation and Training Rewards}
\label{app:domain-trajectories}

Figure~\ref{fig:guru-domain-allocation} tracks the distribution of online rollouts across the six GURU-18K domains at both model scales. Counts include all generated trajectories before filtering, including FEAST's progressive rounds and DAPO's refill attempts. Each point pools counts within a non-overlapping five-update window and is plotted at the window midpoint.

\begin{figure}[!tbp]
\centering
\includegraphics[width=0.92\linewidth]{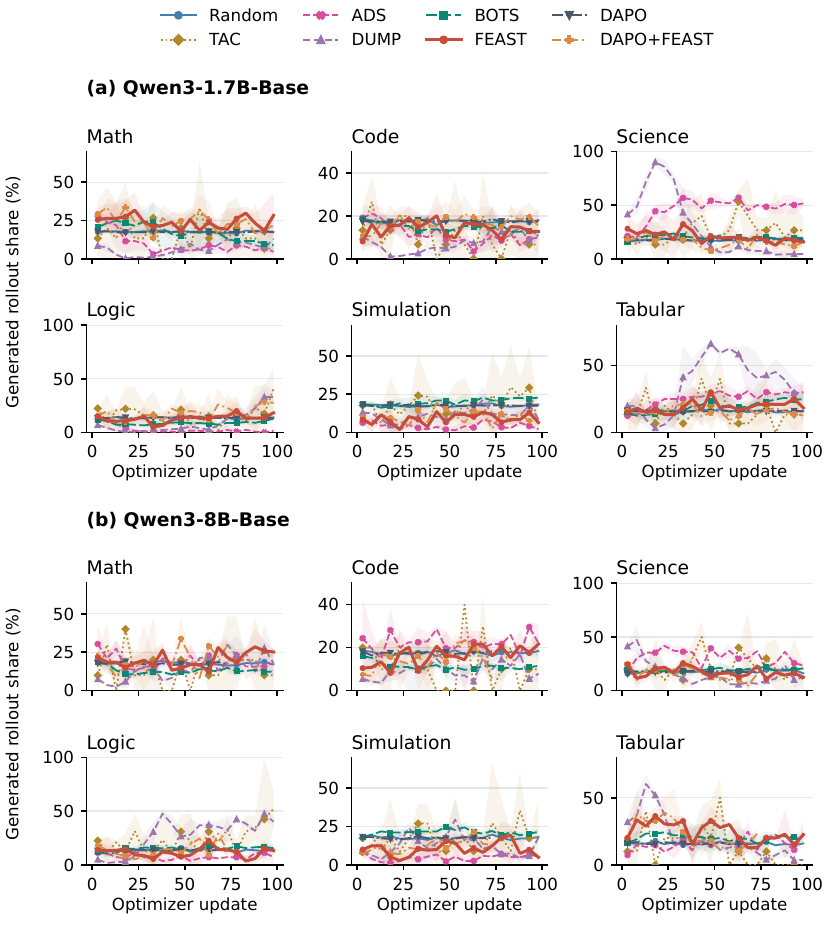}
\caption{Domain-level rollout allocation on GURU-18K. \textbf{(a)} Qwen3-1.7B-Base, seeds 42/43/44. \textbf{(b)} Qwen3-8B-Base, seeds 42/43. Lines show seed means and shaded regions show sample SD. At 8B, DAPO and DAPO+FEAST end at update 75; all other runs end at update 100.}
\label{fig:guru-domain-allocation}
\end{figure}

Figure~\ref{fig:guru-domain-reward} tracks the mean raw verifier reward over generated training trajectories within each domain, retaining partial credit. These rewards depend on the prompts selected by each method and therefore describe training dynamics rather than performance on a common held-out panel.

\begin{figure}[!tbp]
\centering
\includegraphics[width=0.92\linewidth]{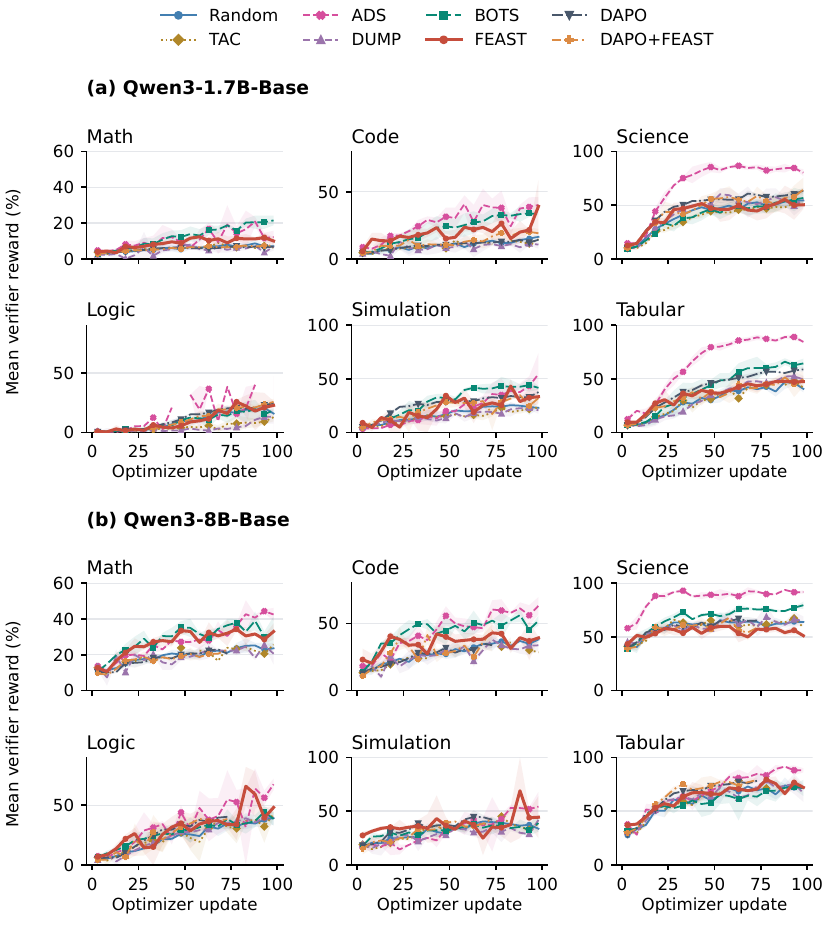}
\caption{Domain-level training rewards on GURU-18K at \textbf{(a)} 1.7B and \textbf{(b)} 8B, with the seeds, windows, and training horizons of Figure~\ref{fig:guru-domain-allocation}. Lines and shaded regions show the mean and sample SD across seeds with domain samples. Windows with no samples from a domain have no reward estimate.}
\label{fig:guru-domain-reward}
\end{figure}
\endgroup

%% file: figures/ratio_allocation_tradeoff.tex
\begin{wrapfigure}{r}{0.42\textwidth}
\centering
\setlength{\abovecaptionskip}{4pt}
\includegraphics[width=\linewidth,trim=0 6bp 0 6bp,clip]{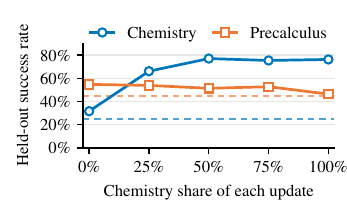}
\caption{Held-out performance after 25 pure-random updates across fixed allocations. Dashed lines indicate the shared step-0 baselines.}
\label{fig:ratio-allocation-tradeoff}
\end{wrapfigure}

%% file: tables/qel_hyperparameters.tex
\begingroup
\fontsize{8}{9}\selectfont
\renewcommand{\arraystretch}{0.96}
\begin{longtable}{@{}>{\raggedright\arraybackslash}p{0.36\linewidth}>{\raggedright\arraybackslash}p{\dimexpr0.64\linewidth-2\tabcolsep\relax}@{}}
\caption{TextArena QEL hyperparameters. S:M:C denotes Sokoban:Mastermind:Countdown.}\label{tab:qel-hyperparameters}\\
\toprule
Hyperparameter & Value \\
\midrule
\endfirsthead
\multicolumn{2}{c}{Table~\thetable{} (continued).} \\
\toprule
Hyperparameter & Value \\
\midrule
\endhead
\midrule
\multicolumn{2}{r}{\textit{Continued on the next page.}} \\
\endfoot
\bottomrule
\endlastfoot
\multicolumn{2}{@{}l}{\textit{Model}} \\*
Base policy & Qwen3-1.7B-Base \\
Adaptation & LoRA on all linear layers; rank 32; $\alpha=32$ \\
Distributed strategy & FSDP; one GPU per training arm \\
\midrule
\noalign{\unskip\nobreak\vskip\belowrulesep}
\multicolumn{2}{@{}l}{\textit{Environment}} \\*
Framework & TextArena \\
Sokoban & Room $6\times6$; one box; maximum 10 turns \\
Mastermind & Code length 3; symbols 1--4; no duplicates; maximum 8 turns \\
Countdown & Six numbers; target absent initially; shortest solution has two operations; maximum 5 turns \\
Policy reward & Strict terminal success: success $=1$, every failure mode $=0$ \\
\midrule
\noalign{\unskip\nobreak\vskip\belowrulesep}
\multicolumn{2}{@{}l}{\textit{Quota schedule}} \\*
S:M:C ratios & $1{:}1{:}1$, $1{:}4{:}8$, $4{:}1{:}8$, $8{:}4{:}1$, $4{:}8{:}1$, $8{:}1{:}4$ \\
Prompt groups per update & 32 \\
Rollouts per prompt group & 8 (256 trajectories per update) \\
Exposure milestones & $n\in\{256,512,1024,2048\}$ selected prompt groups per task \\
Terminal updates & 192 for $1{:}1{:}1$; 832 for each skewed-ratio arm \\
Schedule/data seeds & 1234 / 20260819 \\
\midrule
\noalign{\unskip\nobreak\vskip\belowrulesep}
\multicolumn{2}{@{}l}{\textit{Optimization}} \\*
Objective & GRPO; one PPO epoch \\
Learning rate / schedule & $1\times10^{-5}$ / constant \\
Adam $\beta$ / weight decay & $(0.9,0.999)$ / 0.01 \\
PPO mini-batch size & 8 prompt groups \\
PPO clipping & lower 0.20; upper 0.28 \\
Entropy coefficient & 0.001 \\
KL reward / actor KL loss & disabled / disabled \\
Actor token cap per GPU & 10,240 \\
\midrule
\noalign{\unskip\nobreak\vskip\belowrulesep}
\multicolumn{2}{@{}l}{\textit{Rollout}} \\*
Backend & vLLM server-based asynchronous rollout \\
Sampling & temperature 1.0; top-$p$ 0.95; top-$k$ 20 \\
Per-turn response / thinking budget & 256 / 192 tokens \\
Full trajectory / model context limits & 8,192 / 12,288 tokens \\
Maximum concurrent sequences / batched tokens & 256 / 32,768 \\
\midrule
\noalign{\unskip\nobreak\vskip\belowrulesep}
\multicolumn{2}{@{}l}{\textit{Evaluation}} \\*
Frozen panel & 256 deterministic instances per task; one trajectory per instance \\
Sampling & temperature 0.6; top-$p$ 0.95; top-$k$ 20 \\
Response / thinking / trajectory limits & 256 / 192 / 8,192 tokens \\
Metric & Strict terminal success (\%); compared at matched task exposure $n$ \\
\end{longtable}
\endgroup

%% file: tables/feast_hyperparameters.tex
\begingroup
\fontsize{8}{9}\selectfont
\renewcommand{\arraystretch}{0.96}
\begin{longtable}{@{}>{\raggedright\arraybackslash}p{0.36\linewidth}>{\raggedright\arraybackslash}p{\dimexpr0.64\linewidth-2\tabcolsep\relax}@{}}
\caption{FEAST hyperparameters. Appendix~\ref{app:feast-hyperparameter-rationale} summarizes their theoretical and structural rationale.}\label{tab:feast-hyperparameters}\\
\toprule
Hyperparameter & Value \\
\midrule
\endfirsthead
\multicolumn{2}{c}{Table~\thetable{} (continued).} \\
\toprule
Hyperparameter & Value \\
\midrule
\endhead
\midrule
\multicolumn{2}{r}{\textit{Continued on the next page.}} \\
\endfoot
\bottomrule
\endlastfoot
\multicolumn{2}{@{}l}{\textit{Capability structure}} \\*
Persistent atomic units $J$ & 64; base-policy semantic partition; no task labels \\
Semantic refresh / regroup interval & Current policy / every 10 optimizer updates \\
Regrouping feature & 8-dimensional semantic representation plus smoothed signed progress \\
Dynamic groups & 8--16 diagonal-covariance mixture components; assignment repair to at least 2 atomic units per group \\
Persistent state & Raw attainment, squared pilot differences, learning trend, and rollout shortfall remain attached to atomic units \\
\midrule
\multicolumn{2}{@{}l}{\textit{Inter-cluster fairness}} \\*
Selected groups $M$ / prompts per group & 4 / 32 \\
Sampling & Sequential weighted sampling without replacement; temperature 1.0 \\
Productive signal / attainment & First-round squared raw-reward difference and raw-reward mean; prompt-equal EMA coefficient 0.10 \\
Priority floor / rollout shortfall & Floor $\epsilon=0.05$; shortfall weight $\lambda_a=0.05$; shortfall normalized by 10 equal-share rollout increments and capped at 1 \\
Adaptive surplus & Frontier width $\sigma=0.22$; under-mastered bonus $\eta=0.20$; transition scale $\tau=0.05$; absolute-progress coefficient $\gamma=0.50$ \\
Atomic prompt quota & Lifetime-unseen capacity weighted by atomic priority; eligible-capacity fallback after exhaustion \\
\midrule
\multicolumn{2}{@{}l}{\textit{Intra-cluster efficiency}} \\*
Candidate / selected prompts per group & At most 64 candidates; 32 selected; stable without replacement within each atomic-unit cycle, with repeats only after exhaustion \\
Progressive rollout rounds & $[2,4,16]$ with top-half promotion by cumulative verifier-score variance \\
Final prompt-group sizes per update & 64 groups of 2; 32 groups of 6; 32 groups of 22 \\
Feedback for cluster selection & Attainment and productivity: prompt-equal first-round outcomes; progress: normalized by signal-bearing trajectories from all rounds \\
Optimizer filtering & Prompt groups with zero verifier-score variance do not contribute a reward-centered GRPO update \\
Total rollout budget & 1,024 per update \\
\end{longtable}
\endgroup